\documentclass[10pt,twocolumn,letterpaper]{article}

\usepackage[pagenumbers]{cvpr} 

\definecolor{cvprblue}{rgb}{0.21,0.49,0.74}
\usepackage[pagebackref,breaklinks,colorlinks,allcolors=cvprblue]{hyperref}

\def\paperID{8389} 
\def\confName{CVPR}
\def\confYear{2025}

\newcommand{\comment}[1]{}

\definecolor{colorYes}{RGB}{51,160,44}
\definecolor{colorNo}{RGB}{228,26,28} %

\def\ourmethod{{{\textit{ACID}}}\xspace}
\def\ourmethodwithkd{{{\textit{ACED}}}\xspace}
\def\ourmethodeasy{{{\textit{I-ACID}}}\xspace}
\def\ourmethodhard{{{\textit{H-ACID}}}\xspace}

\newcommand{\zimg}{{z}^{\text{img}}}
\newcommand{\ztxt}{{z}^{\vphantom{\text{img}}\text{txt}}}

\def\imagenetval{{IN-val}}

\usepackage[dvipsnames]{xcolor}

\usepackage{algorithm}
\usepackage{amsfonts}       %
\usepackage{amsmath}
\usepackage{amssymb}
\usepackage{amsthm}
\usepackage{array}
\usepackage{bm}
\usepackage{booktabs} 
\usepackage{tabu}
\usepackage{wrapfig}
\usepackage{caption}
\usepackage{colortbl}
\usepackage{empheq}
\usepackage{enumitem}
\usepackage{etoolbox}
\usepackage{float}
\usepackage{makecell}
\usepackage{mathrsfs}
\usepackage{mdframed}
\usepackage{microtype}      %
\usepackage{multirow}
\usepackage{multicol}
\usepackage{nccmath}
\usepackage{nicefrac}       %
\usepackage[noend]{algorithmic}
\usepackage{pifont}
\usepackage{ragged2e}
\usepackage{rotating}
\usepackage{subcaption}
\usepackage{tabularx}
\usepackage{tikz}
\usetikzlibrary{tikzmark}
\usepackage{times}
\usepackage{url}            %
\usepackage{wrapfig}
\usepackage{color, colortbl}
\usepackage{xspace}
\usepackage{thm-restate}
\usepackage{xfrac}

\newcounter{rowcntr}[table]
\renewcommand{\therowcntr}{\arabic{chapter}.\the\numexpr\arabic{table}+1.\arabic{rowcntr}}

\AtBeginEnvironment{tabular}{\setcounter{rowcntr}{0}}
\newcolumntype{H}{>{\setbox0=\hbox\bgroup}c<{\egroup}@{}}

\newcommand{\PreserveBackslash}[1]{\let\temp=\\#1\let\\=\temp}
\newcolumntype{C}[1]{>{\centering\arraybackslash}m{#1}}
\newcolumntype{R}[1]{>{\raggedleft\arraybackslash}m{#1}}
\newcolumntype{L}[1]{>{\raggedright\arraybackslash}m{#1}}
\usepackage{hyperref}
\usepackage{url}
\usepackage{booktabs}
\usepackage{multirow}
\usepackage{wrapfig}
\usepackage{booktabs,siunitx}

\definecolor{DnCBG}{rgb}{0.9, 0.9, 1.}
\usepackage[capitalize]{cleveref}
\crefname{section}{Sec.}{Secs.}
\Crefname{section}{Section}{Sections}
\Crefname{table}{Table}{Tables}
\crefname{table}{Tab.}{Tabs.}

\title{Active Data Curation Effectively Distills Large-Scale Multimodal Models}

\makeatletter
\newcommand*{\inlineequation}[2][]{%
  \begingroup
    \refstepcounter{equation}%
    \ifx\\#1\\%
    \else
      \label{#1}%
    \fi
    \relpenalty=10000 %
    \binoppenalty=10000 %
    \ensuremath{%
      #2%
    }%
    ~\@eqnnum
  \endgroup
}
\makeatother

\author{
Vishaal Udandarao\thanks{\noindent equal contribution $\dagger$equal supervision\\$\ddagger$work done while interning at Google\\correspondence to: \texttt{vishaal.udandarao@bethgelab.org} or\\ \texttt{nikparth@google.com}}~~$^{3,4\ddagger}$ \quad Nikhil Parthasarathy$^{\textbf{\scriptsize{*}}2}$ \quad Muhammad Ferjad Naeem$^{1}$ \quad Talfan Evans$^{2}$ \\ \quad {Samuel Albanie$^{2}$ \quad Federico Tombari$^{1}$ \quad Yongqin Xian$^{1\dagger}$} \quad Alessio Tonioni$^{1\dagger}$ \quad Olivier J. Hénaff$^{2\dagger}$\\
\vspace{0.1cm}
\normalsize{
$^1$Google
$^2$Google DeepMind
$^3$T\"ubingen AI Center, University of T\"ubingen
$^4$University of Cambridge
}}

\begin{document}
\maketitle
\begin{abstract}
Knowledge distillation (KD) is the de facto standard for compressing large-scale multimodal models into smaller ones. Prior works have explored ever more complex KD strategies involving different objectives, 
teacher-ensembles, and weight inheritance. In this work, we explore an alternative, yet simple approach---\textit{active data curation} as \textit{effective distillation} for contrastive multimodal pretraining. Our simple online batch selection method, {\ourmethod}, outperforms strong KD baselines across various model-, data- and compute-configurations. 
Further, we find such an active 
curation strategy to in fact be complementary to standard KD, and can be effectively combined to train highly performant inference-efficient models. 
Our simple and scalable pretraining framework, {\ourmethodwithkd}, achieves state-of-the-art results across 27 zero-shot classification and 
image-text 
retrieval tasks 
with upto {11}{\%} less inference FLOPs. 
We further demonstrate that {\ourmethodwithkd} yields strong vision-encoders for training generative multimodal models, 
outperforming 
larger vision encoders on image-captioning and visual question-answering tasks.
\end{abstract}    
\section{Introduction}
\label{sec:intro}

Deploying multimodal foundation models~\citep{bommasani2021opportunities} like CLIP~\citep{radford2021learning} on edge devices is challenging due to their high inference costs and memory footprints. This motivates the need for 
smaller, inference-efficient models that retain the performance of their larger counterparts. \textit{Knowledge distillation (KD)}~\citep{hinton2015distilling} is a classic model compression technique---a method for transferring knowledge from a large-scale ``teacher'' model into a smaller ``student'' model, via matching student and teacher logits, features or activations. KD has been extensively deployed for creating small, performant models like Gemma-2~\citep{team2024gemma}, Phi-3~\citep{abdin2024phi}, Gemini-1.5 Flash~\citep{reid2024gemini}, and SD3-Turbo~\citep{sauer2024fast}.

\begin{figure}[ht]
    \centering
\includegraphics[width=\linewidth]{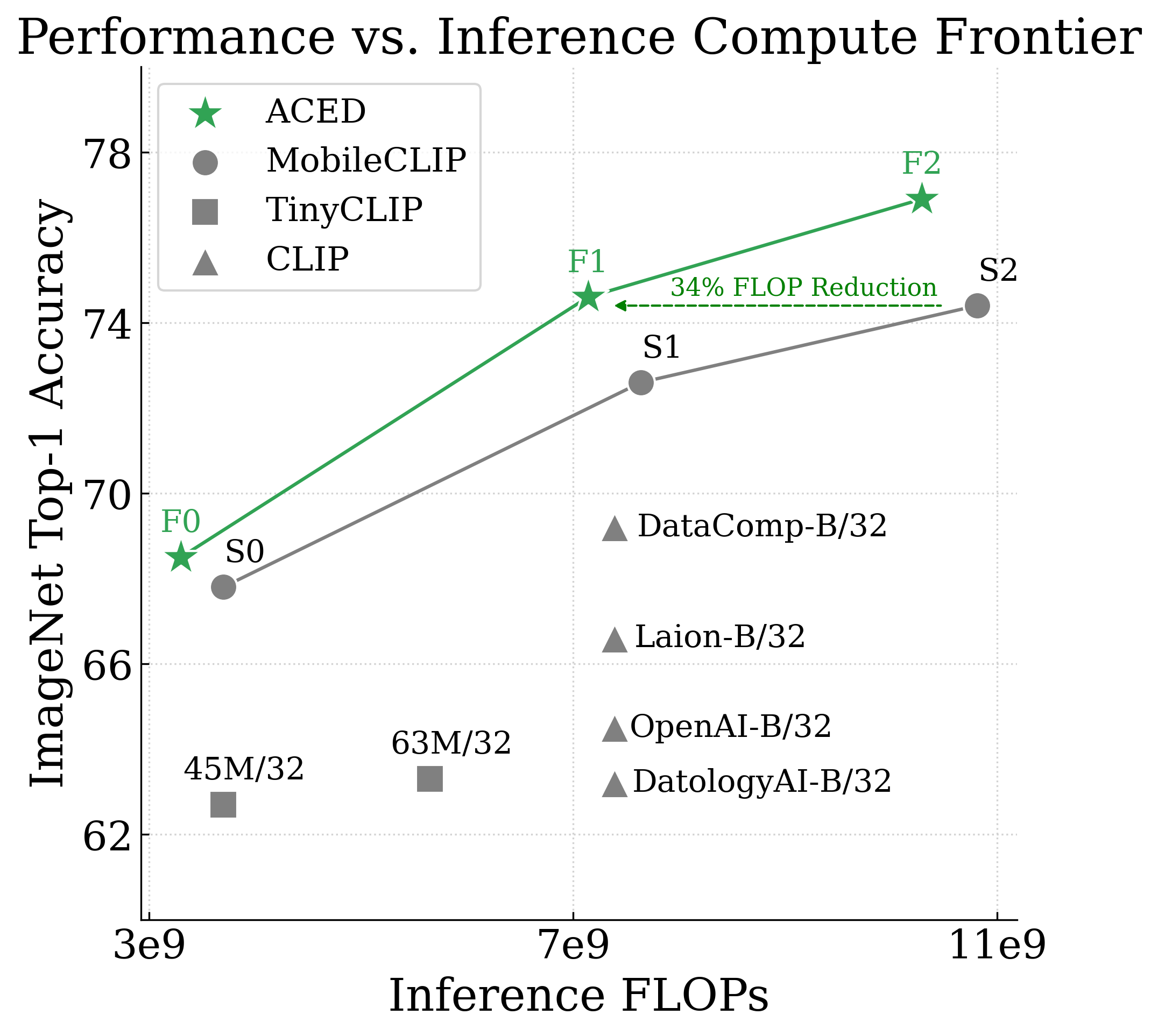}
\caption{\textbf{Performance-Inference Frontier.} Our \ourmethodwithkd models (\textbf{A}ctive \textbf{C}uration with \textbf{E}xplicit \textbf{D}istillation, see~\cref{sec:methods}), achieve a new pareto frontier for performance (measured by ImageNet top-1 zero-shot validation accuracy) vs. inference GFLOPs.}
     \label{fig:aced_teaser}
 \vspace{-8pt}
\end{figure}

Here, our primary goal is to downscale contrastive vision-language
models
, without compromising downstream performance. Prior works in this domain focus on complex KD strategies as the key solution---the current SoTA (TinyCLIP~\citep{wu2023tinyclip} and MobileCLIP~\citep{vasu2024mobileclip}) use combinations of methods such as strong data-augmentation policies, multi-teacher ensembles, synthetic captions, weight-inheritance,  weight-pruning, and bespoke model architectures.

In this work, we seek a simplified approach. Specifically, we propose using \textit{active data curation as an effective strategy for distilling large vision-language models (VLMs)} into smaller and more FLOP-efficient multimodal models. 


 Our method, \ourmethod (\underline{A}ctive \underline{C}uration as \underline{I}mplicit \underline{D}istillation), automatically selects samples that reduce the performance gap between a small \textit{student} model and a larger \textit{reference} model. 
Under appropriate conditions, we find this is a surprisingly effective distillation approach. To the best of our knowledge, this is a novel finding since prior works in data curation assume that larger models can not be used to select data for smaller ones, due to the \textit{capacity gap}~\citep{mindermann2022prioritized,gadre2024datacomp}.
Through a novel theoretical interpretation and extensive experiments, instead, we demonstrate that \ourmethod is not only effective but also improves over standard KD, exhibiting 
more favourable scaling 
with respect to training compute. 
We also conduct careful ablation studies that uncover factors influencing the quality of the trained student model, including, reference model capacity and 
training dataset.

After comprehensively demonstrating the effectiveness of data curation as an alternative to KD
, we further show how the two can be 
profitably
combined to further improve performance. 
This suggests that the \textit{information distilled to the smaller model through each approach is complementary}.

Based on this finding, we propose our final pretraining recipe, \textbf{\textit{ACED}} (\underline{AC}ID with \underline{E}xplicit \underline{D}istillation), andtrain very strong FLOP-efficient image-text contrastive models. Our method, absent bespoke components such as efficient architectures or data-augmentations, outperforms SoTA CLIP and SigLIP models with greater FLOP-efficiency at inference-time and shows a significant improvement over 27 downstream tasks against these prior SoTA FLOP-efficient models~\citep{vasu2024mobileclip,wu2023tinyclip}. We further demonstrate that our \ourmethodwithkd vision-encoders provide strong backbones for generative multimodal models, outperforming larger and FLOP-inefficient vision-encoders on image-captioning and visual-question-answering (VQA) tasks.

\section{Related Work}
\label{sec:relwork}

\noindent\textbf{Multimodal Data Curation.} 
Recent works have emphasised the importance of data quality for multimodal pretraining~\citep{nguyen2022quality, gadre2024datacomp,udandarao2024no,fang2022data,mayilvahanan2023does}.
Specifically, offline curation of noisy web-scale data can result in large pretraining efficiency gains \citep{jia2021scaling,changpinyo2021conceptual,wang2023too,abbas2023semdedup,cao2023less,maini2023t, xu2023demystifying,abbas2024effective, mahmoud2024sieve,wang2024variance,fang2023data}.
However, these static methods that pre-filter data do not take into account the training dynamics of the current learner model.
As a result, there have been many recent attempts to introduce \textit{online batch selection} criteria that account for the current state of the learner (\textit{e.g.}, at each step select training samples that have the largest learner loss) \citep{loshchilov2015online, ioannou2023online,joseph2019submodular,song2020carpe,jiang2019accelerating,wang2024efficienttrain,zhou2024multi,schaul2015prioritized,xu2023cit}.
The RHO-Loss~\citep{mindermann2022prioritized} goes further to consider current learner state and a pretrained data-selector (reference) model. This criterion has since been used in many efforts to improve the efficiency of foundation model pretraining~\citep{evans2023bad,evans2024data,brandfonbrener2024color,hong2024diversified,fan2023irreducible,deng2023towards}.
As these methods seek to improve pretraining efficiency, the pretrained reference models that are used as \textit{data selectors are typically smaller than the learner models they are used to train}~\citep{evans2023bad,evans2024data,fang2023data}. In fact, many works have shown that increasing reference model size can potentially hurt learner model performance~\citep{fang2023data,yu2023devil,gadre2024datacomp} . Our work tells a different story, finding that \textit{large data selectors can effectively curate data for inference-time FLOP-efficient learner models}.



\noindent\textbf{Knowledge Distillation.} First introduced by~\citet{buciluǎ2006model} and further popularized by~\citet{hinton2015distilling,ba2014deep}, knowledge distillation (KD) is a classic technique for transferring knowledge from a larger model (\textit{teacher}) to a smaller one (\textit{student}) by optimizing the student to match outputs of the teacher. Such methods have been used for compressing large models in unimodal tasks like image-classification~\citep{beyer2022knowledge,cho2019efficacy,vemulapalliknowledge,wang2020knowledge,nix2023hard,tian2019contrastive} and language representation learning~\citep{xu2024survey,agarwal2024policy,sanh2019distilbert,kim2016sequence,lin2020autoregressive,hahn2019self,tan2023gkd}. Further works have extended KD to use teacher-ensembles~\citep{shen2020meal,chebotar2016distilling,you2017learning,stanton2021does,malinin2019ensemble,faghri2023reinforce,zuchniak2023multi,sariyildiz2024unic}, and different distillation training objectives~\citep{ji2021show,zhao2024no,xie2020self,tarvainen2017mean,touvron2021training,ranzinger2024radio,li2024promptkd}. 

Most relevant to our work, 
 there are a number of recent efforts to distill CLIP 
models.
SF-CLIP~\citep{sameni2024building} explores masked distillation, while MobileCLIP~\citep{vasu2024mobileclip} uses multi-teacher contrastive-KD, synthetic captions, and data-augmentations. TinyCLIP~\citep{wu2023tinyclip} proposes a weight inheritance method combined with an affinity-mimicking strategy. 
An empirical study (CLIP-KD~\citep{yang2023clip}) also has explored different objective functions for effectively distilling CLIP models, across different scales.
Finally, CLIP-CID~\citep{yang2024clip} uses an image semantic balancing strategy coupled with cluster-instance discrimination for better teacher-to-student knowledge transfer during the KD process.
We compare against all of these methods in our experimental results in~\cref{sec:experiments}.

\noindent \textbf{Accelerating Knowledge Distillation.} 
Prior works have investigated accelerating vanilla KD using active learning in small-scale settings~\citep{wang2020neural,lan2024improve,xu2023computation}. However, these approaches require a costly iterative process, involving synthetic generation, followed by active sample selection to produce pseudo-labels from a teacher model, thereby limiting their scalability. 
Other works have studied data-selection methods for improving KD, typically using uncertainty-based data, logit and feature selection~\citep{rao2023dynamic,roth2023fantastic,wang2023improved,wang2022multimodal,lin2022efficient,zhou2023adads,li2021dynamic,he2022knowledge,wang2024cascade}, contextual retrieval and sample augmentation from a large data pool~\citep{liu2022rethinking,jiao2019tinybert,liang2020mixkd,ge2024training,zhang2023reaugkd,radenovic2023filtering}, or influence-function based sample selection~\citep{lan2024improve,ye2022progen}. 
Contrary to these works, others suggest that vanilla knowledge distillation is optimal in ``infinite-data regimes''~\citep{beyer2022knowledge, hao2024revisit}. Surprisingly, these studies operate primarily in the unimodal image/text classification regime, and none have been scaled to multimodal foundation model training. 

We showcase, for the first time, that \textit{simple data selection using online batch selection outperforms standard KD for pretraining multimodal models}. We further study the optimal strategies for combining vanilla KD and active data curation in order to best leverage their complementary strengths.

\section{Methods}
\label{sec:methods}

\subsection{Preliminaries}
\label{sec:preliminaries}

 
\noindent\textbf{Contrastive Vision-Language Pretraining.} We follow standard multimodal pretraining frameworks like CLIP~\citep{radford2021learning} and SigLIP~\citep{zhai2023sigmoid}. 
We assume a large pretraining dataset $\mathcal{D}$, containing image-text pairs.
Our goal is to train a two-tower VLM with parameters ${\theta}$ whose image-encoder ${f}^{\text{img}}$ and text-encoder ${f}^{\text{txt}}$ are initialized from scratch.
At each training step, we sample a mini-batch, $\mathcal{B}{=}\{{x_1,\dots,x_b}\}$, where $x_i = (I_i,T_i)$ denotes the $i^{\text{th}}$ image-text pair in the mini-batch and $b$ denotes the batch-size.
We then encode and normalize the embeddings of each image-text pair in the mini-batch as ${z}^{\text{img}}_{i}{=}{\frac{{{f}^{\text{img}}{({I_i}{|}{\theta})}}}{\| {{f}^{\text{img}}{({I_i}{|}{\theta})}} \|_{2}}}$ and ${z}^{\text{txt}}_{i}{=}{\frac{{{f}^{\text{txt}}{({T_i}{|}{\theta})}}}{\| {{f}^{\text{txt}}{({T_i}{|}{\theta})}} \|_{2}}}$. The pairwise similarities $l_{ij}(\theta) = {\alpha}{\zimg_{i}}{\cdot} {\ztxt_{j} + \beta}$, where $\alpha, \beta$ are learnable inverse-temperature and offset hyperparameters, 
can be converted into pairwise probabilities with a row- or column-wise softmax as follows, 
\begin{align}
\label{eq:probs}
p^{\text{img} \rightarrow \text{txt}}_{ij} & = \exp(l_{ij}) / \sum_{{k}{=}{1}}^{b}\exp(l_{ik}) \\
p^{\text{txt} \rightarrow \text{img}}_{ij} & = \exp(l_{ij}) / \sum_{{k}{=}{1}}^{b}\exp(l_{ki})\label{probs2}
\end{align}
or $p^{\text{sig}}_{ij} = \sigma(l_{ij})$ with a sigmoid operation.
The contrastive image-text losses align embeddings of paired images and texts
$(\zimg_{i},\ztxt_{i})$, while pushing apart embeddings of mismatched images and texts $(\zimg_{i},\ztxt_{{j}{\neq}{i}})$. There are two widely used contrastive variants i.e., 
$\mathcal{L}_{\text{softmax}}$ for CLIP~\cite{radford2021learning} and $\mathcal{L}_{\text{sigmoid}}$ for SigLIP~\citep{zhai2023sigmoid},
both of which can be framed as ${\mathcal{L}(x_i; \mathcal{B})}{=}{-}{\sum_{j=1}^b y_j(x_i) \log p_{ij}}{=}{\text{CE}[ y(x_i); p(x_i)]}$ for a suitable choice of binary labels $y$ and probabilities $p$, where CE is the standard cross-entropy loss (see \cref{contrastive-loss-appendix} for details). 
 By default, we use the sigmoid variant as it is more scalable, 
but also run ablations with the softmax variant.


\vspace{0.5em} \noindent\textbf{Contrastive Distillation.} 
Given the student $\theta$ and a pretrained teacher model $\theta_{\text{teacher}}$, our aim is to distill the contrastive logit matrix from teacher to student. Formally, given a data-batch $\mathcal{B}$, we extract teacher embeddings $(\zimg_{i},\ztxt_{i})(\theta_\text{teacher})$ and student embeddings $(\zimg_{i},\ztxt_{i})(\theta)$, yielding pairwise similarities $l_{ij}(\theta_\text{teacher})$ and $l_{ij}(\theta)$ for the teacher and student respectively. Let $p$ and $q$ be the pairwise probabilities induced by teacher and student similarities (\cref{eq:probs,probs2}). 
Our knowledge distillation (KD) objective is simply the cross-entropy loss between these distributions: %
\begin{multline}
\label{eq:smax-distillation-loss}
\mathcal{L}_{\text{dist}}(x_i; \mathcal{B}) = \text{KD}[p(x_i), q(x_i)] = \\
-\frac{1}{2} \sum_{j=1}^{b} \left( p^{\text{img} \rightarrow \text{txt}}_{i,j} \log q^{\text{img} \rightarrow \text{txt}}_{i,j} + p^{\text{txt} \rightarrow \text{img}}_{i,j} \log q^{\text{txt} \rightarrow \text{img}}_{i,j} \right)
\end{multline}
which has previously been  explored 
in unimodal~\citep{yang2023online,fang2021seed} 
and multimodal contexts~\citep{yang2024clip}. 


\begin{figure*}[t]
    \centering
    \includegraphics[width=0.9\linewidth]{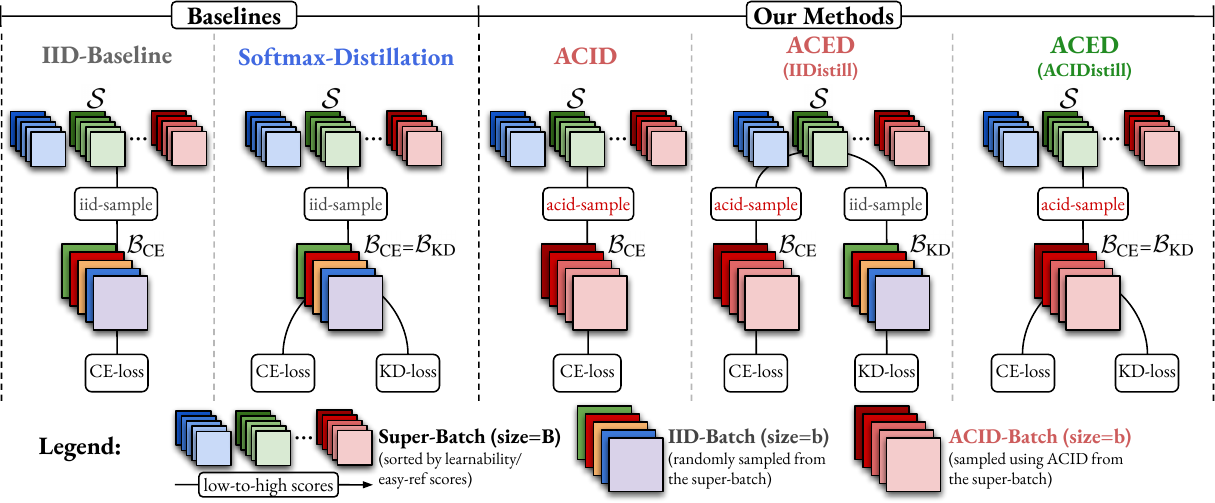}
    \caption{\textbf{Different Method Configurations.} We depict all the different method configurations that we consider in our work. Each method can be independently recovered from the unified objective $\mathcal{L}_\text{full}$ in~\cref{combination-section}. The \texttt{\textcolor{Gray}{\textbf{iid-sample}}} and \texttt{\textcolor{Maroon}{\textbf{acid-sample}}} boxes denote the IID-sampling and our \textit{ACID} online batch-selection sampling schemes respectively. For more details, refer to~\cref{sec:methods}.}
    \label{fig:methods-figure}
\vspace{-1em}
\end{figure*}
\subsection{\textbf{\textit{ACID}}: \underline{A}ctive \underline{C}uration as 
\underline{I}mplicit \underline{D}istillation}
\label{sec:selectdist}

\noindent\textbf{Setup.} We refer to the small model we aim to train
as the \textit{student} model, with parameters $\theta$. Given an image-text pretraining dataset $\mathcal{D}$, the straightforward training approach is to sample uniformly random batches of data $\mathcal{B}$ (of size $b$), from $\mathcal{D}$ at each step $t$, and minimize $\mathcal{L} \in \{ \mathcal{L}_{\text{softmax}}, \mathcal{L}_{\text{sigmoid}}\}$. We refer to this baseline strategy, minimizing $\hat{\mathcal{L}} = \frac{1}{b} \sum_{x_i \sim \mathcal{U}[\mathcal{D}]} \mathcal{L}(x_i; \mathcal{B})$ as the \textit{IID-baseline} ($\theta_{\text{IID}}$) 

\vspace{0.5em} \noindent\textbf{Active Data Curation} employs a smarter way to select batches, using a pretrained \textit{reference} model $\theta_{\text{ref}}$. At each step $t$, we select a sub-batch $\mathcal{B}$ (size $b$) from a much larger super-batch $\mathcal{S}$ (size $B$) according to an \textit{active selection distribution} $\mathcal{A}[\mathcal{S}]$. We use two main criteria for scoring sub-batches $\mathcal{B}$, following prior work in prioritized sampling~\citep{mindermann2022prioritized,evans2023bad}. 
\begin{enumerate}
    \item \textit{Easy-reference scoring} uses the loss-values of the reference $\theta_{\text{ref}}$ 
    to preferentially sample batches that are easy for $\theta_{\text{ref}}$: ${s^{\text{easy\_ref}}(\mathcal{
B}|\theta_{\text{ref}})}{=}{-\mathcal{L}(\mathcal{B}|\theta_{\text{ref}})}$.
    \item  \textit{Learnability scoring} uses the difference in loss-values of the current student $\theta$ and the reference $\theta_{\text{ref}}$ 
    to give high scores to \textit{learnable batches} i.e., batches that are easy for the reference but difficult for the current student: ${s^{\text{learn}}(\mathcal{
B}|\theta,\theta_{\text{ref}})}{=}{{\mathcal{L}(\mathcal{B}|\theta)}{-}{\mathcal{L}(\mathcal{B}|\theta_{\text{ref}})}}$. 
\end{enumerate}
Prior model-based online batch curation methods used reference models that were \textit{of the same size or smaller} than the model being trained. This was because of (1) \textit{training efficiency}: since data-selection was originally used to reduce training set sizes, reference models were 
chosen to be small so as to reduce
compute
overhead, and (2) \textit{unlearnable prioritization}: intuitively, samples that are easily learned (and thus prioritized) by a high-capacity reference might be unlearnable for the lower-capacity learner. 
Indeed \citet{mindermann2022prioritized} observed little effect when increasing reference model capacity, a key limitation of their original method. 

\vspace{0.5em} \noindent\textbf{\underline{A}ctive Data \underline{C}uration as \underline{I}mplicit \underline{D}istillation (ACID).} We now show formally that 
active curation can be cast as ``implicit distillation'' and should benefit from larger reference models. 
The model now minimizes 
$\hat{\mathcal{L}} = \frac{1}{b} \sum_{x_i \sim \mathcal{A}[\mathcal{S}]} \mathcal{L}(x_i; \mathcal{B})$, which in expectation is $ \mathcal{E} = \mathbb{E}[\hat{\mathcal{L}}] = \sum_{x\in\mathcal{D}} a(x) \mathcal{L}(x; \mathcal{B})$ given that super-batches $\mathcal{S}$ are sampled uniformly.
Recall that $\mathcal{L}(x; \mathcal{B}) = -\sum_{i=1}^b y_i(x) \log q_i(x)$
, where $y_i$ are the labels of the contrastive task and $q_i$ are the probabilities induced by the pairwise similarities of the student $\theta$. Let $p_i$ be the probabilities induced by the reference model $\theta_{\text{ref}}$. In the case of \textit{easy-reference scoring} and the softmax loss, $a(x) = \frac{1}{Z} \exp \sum_{i=1}^b y_i(x) \log p_i(x) = \frac{1}{Z} p_{i^*}(x)$ where $i^*$ is the index of the one-hot label $y(x)$. We derive the following equality (see \cref{acid-derivation-appendix} for details), 
\begin{align}
\label{eq:acid}
    \mathcal{E}_\text{easy-ref} & = \frac{1}{Z} \sum_{x \in \mathcal{D}} \text{KD}[ p(x) \cdot y(x) ; q(x) ].
\end{align}
This demonstrates that by curating data according to the reference model $\theta_\text{ref}$, we implicitly distill its knowledge via a novel data-driven objective, using a combination of model predictions and real labels as targets. Model predictions and real labels have independent sources of noise: false labels can occur due to human error, whereas models may underfit due to biases in training or architecture. As a result, retaining targets where the reference model and labels agree allows for mutual denoising of model predictions and data labels. 

Moreover, this suggests that in contrast to the standard active learning paradigm, in which reference models are similarly-sized or smaller than the student model~\citep{mindermann2022prioritized, evans2023bad},  
\ourmethod should instead benefit from pretrained reference models $\theta_{\text{ref}}$ that are \textit{larger} than the student model $\theta$ for scoring. While counter-intuitive from an active learning perspective, this configuration is natural given our new perspective of active data curation as an implicit form of distillation. 


\vspace{0.5em}\noindent\textbf{Learnability-based Data Curation is Hard Distillation.} When using learnability-based prioritization, the active selection distribution $\mathcal{A}$ factorizes as 
$a^\text{learn} = \frac{1}{Z} \exp(s^\text{learn}) = \frac{1}{Z} \exp[\mathcal{L}(\cdot | \theta) - \mathcal{L}(\cdot | \theta_\text{ref})] = a^\text{easy-ref} \cdot a^\text{hard-learn}$ where $a^\text{hard-learn} = \frac{1}{Z} \exp[\mathcal{L}(\cdot | \theta)]$ prioritizes examples with high loss according to the student.
Since easy-reference prioritization yields implicit distillation (\ourmethodeasy, \cref{eq:acid}), learnability prioritization yields
\begin{align}    
    \mathcal{E}_\text{learn} & = \frac{1}{Z} \sum_{x \in \mathcal{D}}  a^\text{hard-learn}(x) \text{KD}[ p(x) \cdot y(x) ; q(x) ] 
\end{align}
i.e.\ implicit distillation on hard examples (``\ourmethodhard'') according to the student (see \cref{acid-derivation-appendix} for details). Prioritizing high-loss examples has been shown to reliably accelerate learning in settings where targets are high-quality~\citep{loshchilov2015online}, as is the case with the combined targets in our ACID.


\vspace{0.5em} \noindent\textbf{Joint  Batch Sampling.}
Implementing our ACID method requires sampling examples $x$ from $\mathcal{A}[\mathcal{S}]$
where ${a(x|\mathcal{B})}{=}{\exp({-}{\mathcal{L}(x|\mathcal{B},\theta_{\text{ref}})})}$ for ACID or $a(x|\mathcal{B}) = \exp({\mathcal{L}(x|\mathcal{B},\theta)}{-}{\mathcal{L}(x|\mathcal{B},\theta_{\text{ref}})})$ for Hard-ACID. As such, sampling from $\mathcal{A}[\mathcal{S}]$ requires \textit{jointly selecting examples in a batch}. Following \citet{evans2024data} we utilise an iterative approach which incrementally populates the batch conditioned on already-sampled examples.
Specifically, this algorithm uses $n$ iterations of a blocked Gibbs sampling approach. Given a subset of data-samples $\mathcal{B}_{i}$ at iteration $i$, we compute the conditional batch-scores of all other candidate samples in the super-batch that have not yet been added to the mini-batch $\mathcal{B}_i$, $s^{\text{easy\_ref}}({\{\mathcal{B}_i,x\}})$/$s^{\text{learn}}({\{\mathcal{B}_i,x\}}){\forall}{{x}{\in}{{\mathcal{S}}{-}{\mathcal{B}_i}}}$, then sample a chunk $\{x_k\}$ of size $\frac{b}{n}$ according to these scores independently, and append to the constructed mini-batch, $\mathcal{B}_{{i}{+}{1}}{=}{\mathcal{B}_{i}}{\cup}{\{x_k\}}$. The first chunk $\mathcal{B}_{1}$ is sampled using the independent scores $s^{\text{easy\_ref}}(\{x\})$/$s^{\text{learn}}(\{x\})$. The final sampled mini-batch is yielded after $n$ iterations, $\mathcal{B}{=}{\mathcal{B}_{{n}}}$ (see \citet{evans2024data} for more details). Note that the ratio of the super-batch size and the mini-batch size determines how aggressively our data selection method filters out samples from the super-batch---we quantify this with the filtering ratio, ${f}{=}{{1}{-}{\frac{b}{B}}}$. The larger the filtering ratio $f$, the stronger is the data selection process at each training step. 

\begin{table}[h]
\centering
\resizebox{0.99\linewidth}{!}{
\begin{tabular}{c|ccc|c} 
\toprule
    \textbf{Method} & 
    {$\mathbf{\mathcal{\lambda}}$} &
    {$\mathbf{\mathcal{B}_{\text{CE}}}$} & {$\mathbf{\mathcal{B}_{\text{KD}}}$} &
    {\textbf{Effective Batch-Size per Iteration}}
    \\
    \midrule
    \textcolor{Gray}{\textit{IID-Baseline}} & ${=}{0}$ & IID & --- & $b$ \\
    \textcolor{RoyalBlue}{\textit{Softmax-KD}} & ${>}{0}$ &IID & IID  & $b$ \\
    \midrule
    \textcolor{Maroon}{\textit{\ourmethodeasy}} & ${=}{0}$ & \ourmethodeasy & --- & $b$ \\
    \textcolor{Maroon}{\textit{\ourmethodhard}} & ${=}{0}$ & \ourmethodhard & --- & $b$ \\ 
    \textcolor{Maroon}{\textit{ACED-IIDistill}} & ${>}{0}$ & \ourmethodhard & IID & $2b$ \\ 
    \textcolor{ForestGreen}{\textit{ACED-ACIDistill}} & ${>}{0}$ & \ourmethodhard & \ourmethodhard & $b$ \\ 
\bottomrule
\end{tabular}}
\caption{\textbf{Method Instantiations} recovered from our unified objective (see~\cref{combination-section}), by specifying data-selection strategies across different batches and hyperparameter values. We further indicate the effective mini-batch size per-iteration used by each method, and colour-code different methods for easy referencing from~\cref{sec:experiments}.}
\label{tab:method-instantiations}
\vspace{-2mm}
\end{table}

\subsection{\textbf{\ourmethodwithkd}: \underline{A}ctive \underline{C}uration \& \underline{E}xplicit \underline{D}istillation}
\label{combination-section}

\noindent\textbf{Towards explicit knowledge-transfer.} {ACID} introduces an active curation strategy without using any auxiliary objective beyond the contrastive loss. This induces an implicit form of knowledge transfer from the larger reference model to the small student model. 
To augment this implicit transfer with an explicit distillation objective, we  propose \ourmethodwithkd, \underline{{AC}}{ID} with \underline{E}xplicit \underline{D}isillation, which effectively combines {ACID} with a softmax contrastive distillation loss (see~\cref{eq:smax-distillation-loss}). 


\noindent\textbf{A unified objective.} 
We now propose a general loss formulation that can flexibly model different instantiations of all our training methods (\textit{IID-Baseline}, \textit{\ourmethod}, \textit{\ourmethodwithkd}, and \textit{Softmax-KD}) under one unified objective. 
At each step $t$, we first sample the super-batch $\mathcal{S}$
based on the required final mini-batch size $b$ and filtering ratio $f$ (super-batch size is ${B}{=}{\frac{{b}}{{1}{-}{f}}}$).
We then sample two mini-batches from $\mathcal{S}$---the data mini-batch used for training the contrastive loss ($\mathcal{B}_{\text{CE}}$) and the mini-batch used for distillation ($\mathcal{B}_{\text{KD}}$). 
The two mini-batches can either be sampled using our \textit{ACID} sampling scheme or random IID sampling.
Our overall objective is written as, $\mathcal{L}_\text{full} = \mathcal{L}_{\text{softmax/sigmoid}}[\mathcal{B}_{\text{CE}}] + \lambda\cdot\mathcal{L}_{\text{dist}}[\mathcal{B}_{\text{KD}}]$.

\cref{tab:method-instantiations} and~\cref{fig:methods-figure} depict how we can instantiate $\mathcal{L}_\text{full}$ to recover different methods and baselines---we colour-code different methods to enable easy cross-referencing later from~\cref{sec:experiments}.
Our \textcolor{Gray}{\textit{IID-Baseline}} only uses the contrastive loss trained on an IID-sampled batch. Our implicit distillation methods (\textcolor{Maroon}{\textit{\{I/H\}-\ourmethod}}) also use only the contrastive loss but train on actively selected data-batches. For \textcolor{RoyalBlue}{\textit{Softmax-KD}}, we only sample an IID batch and use that same batch for both contrastive and distillation losses (${\mathcal{B}_{\text{CE}}}{=}{\mathcal{B}_{\text{dist}}}$). For our combined \textcolor{ForestGreen}{\textit{\ourmethodwithkd}} method, we have two schemes---(1) \textcolor{ForestGreen}{\textit{ACIDistill}} which samples a single mini-batch from $\mathcal{S}$ using \textcolor{Maroon}{\ourmethodhard}, using that for both contrastive and distillation training (${\mathcal{B}_{\text{CE}}}{=}{\mathcal{B}_{\text{KD}}}$), and (2) \textcolor{Maroon}{\textit{IIDistill}} which samples $\mathcal{B}_{\text{CE}}$ using \textcolor{Maroon}{\ourmethodhard} and $\mathcal{B}_{\text{KD}}$ using \textcolor{Gray}{IID} sampling. For both \textcolor{ForestGreen}{\textit{\ourmethodwithkd}} methods, we only use the \textcolor{Maroon}{\ourmethodhard} sampling scheme as empirically it is more performant than \textcolor{Maroon}{\ourmethodeasy} (see~\cref{fig1}).






\section{Experiments}
\label{sec:experiments}

\subsection{Implementation Details}
\label{sec:impdetails}
\noindent\textbf{Model Architecture and Sizes.} Unless otherwise specified, we use standard ViT-S~\citep{dosovitskiy2020image} and BERT-small~\citep{devlin2018bert} models 
as our student image-text encoders.
For some student ablations, we also use (ViT-Ti image, Ti text) and (ViT-B image, B text) configurations.  
For our references and teachers, we sweep over different 
sizes---(ViT-Ti, Ti), (ViT-S, S), (ViT-B, B), (ViT-L, L), (ViT-H, H), and (ViT-g, g) for (image, text) encoders respectively. We pretrain all our models ($\theta_{\text{teacher}}$, $\theta_{\text{ref}}$, $\theta$) from scratch. For more details, refer to \cref{expansion-on-model-details}.

\vspace{0.5em} \noindent\textbf{Pretraining Datasets.} We use the popular DataComp-1B~\citep{gadre2024datacomp} dataset for pretraining all our student models. For training our reference and teacher models, we sweep over four different datasets---WebLI-curated++~\citep{evans2024data}, WebLI-1B~\citep{chen2022pali}, LAION-400M~\citep{schuhmann2021laion}, and DataComp-1B~\citep{gadre2024datacomp}.

\begin{figure}
\centering
\includegraphics[width=\linewidth]{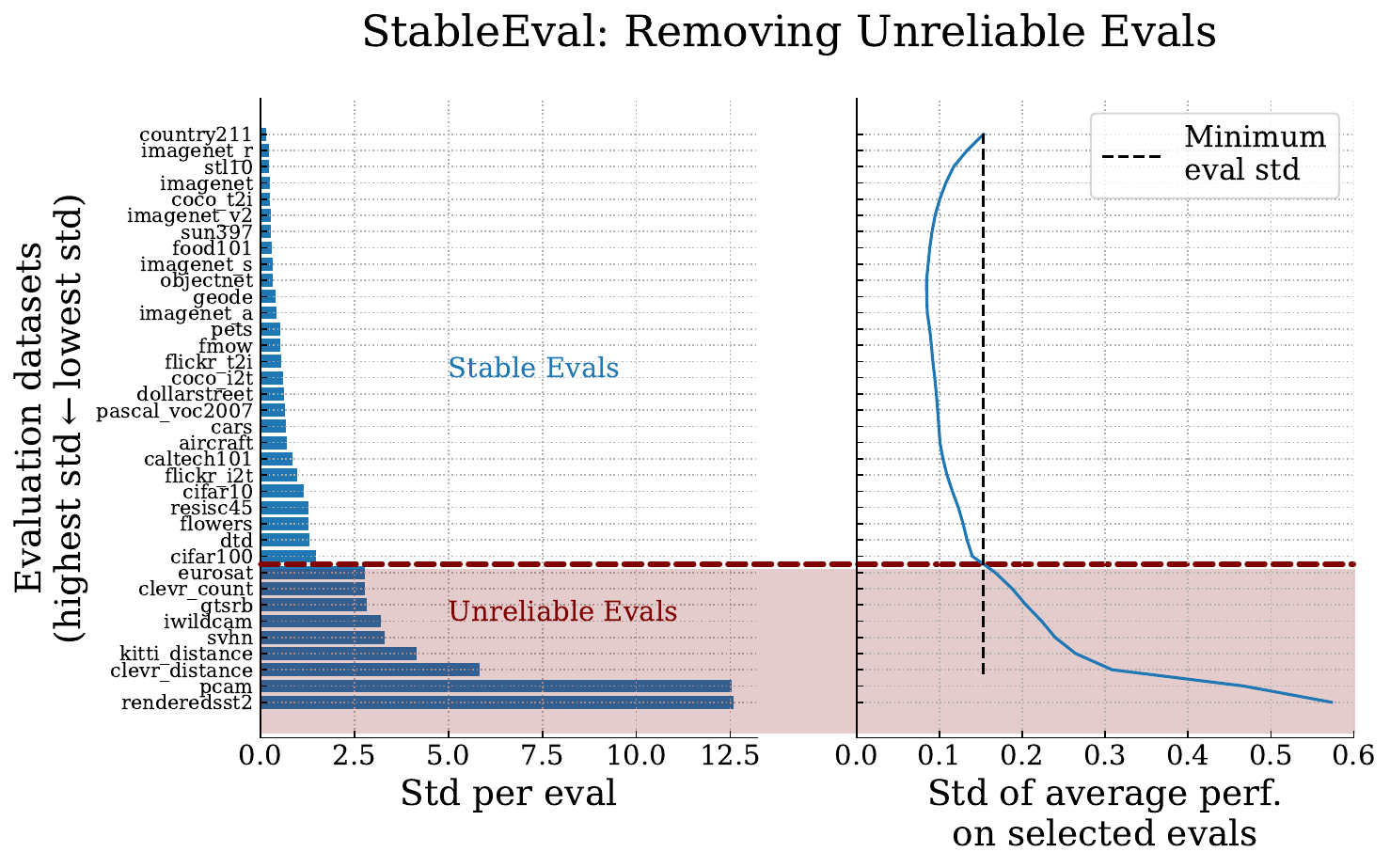}
\vspace{-15pt}
\caption{
\textbf{StableEval: a reliable set of multimodal evaluations.} (\textit{left}) Variability across random pretraining seeds of individual evaluations. (\textit{right}) Variability of average performance across incrementally larger sets of evaluations, starting from the most reliable.  \vspace{-1em}
}
\label{fig:stable-eval}
\end{figure}

\vspace{0.5em} \noindent\textbf{Evaluation Protocol: StableEval.} We evaluate our models on a diverse set of benchmarks including zero-shot classification and image-text retrieval datasets following prior multimodal pretraining works~\citep{gadre2024datacomp,lavoie2024modeling,xu2023demystifying}. 
However, many works select non-standardized sets of evaluations and fail to sufficiently justify the reliability of the evaluations they use.
To rigorously define an evaluation suite, we collate a standard list of 34 candidate evaluations and conduct a systematic analysis of their reliability. By repeating the same canonical pretraining run multiple times (\textit{e.g.},\ CLIP pretraining on DataComp with the \textit{exact same data ordering}, see \cref{appx-sec:evaluation} for details), we evaluate the variability of each metric across random seeds. In \cref{fig:stable-eval} (left), we find an extreme range in variability across evaluations (stds from 0.15\% to 12.5\%) which hinders comparisons among different methods. Inspired loosely by the continuous inverse-variance weighting (IVW) method for minimizing variance of aggregated random variables \cite{hartung2011statistical}, we develop a method for choosing a discrete, stable subset of relevant evaluations. We compute the variability of a progressively growing set of evaluations, starting from least variable and incrementally adding more variable ones, in ascending order. For a subset of size $N$, ${std}{(E_1 ... E_N)}{=}{\sqrt{\frac{1}{N^2}\sum_i{var(E_i)}}}$.
Because of the $1/N^2$ scaling, adding more datasets decreases the variability of the average (\cref{fig:stable-eval} (right)) to a critical point. However, adding highly variable evaluations outweighs this term, increasing the average variability. We limit the evaluation set to remain highly reliable (i.e.\ with lower variability than the most reliable individual evaluation (${<}{0.15}$)) while still including as many evaluations possible to maximize coverage and diversity, yielding the 27 {\textit{StableEval}} set.

\begin{figure} 
\centering
    \centering
    \includegraphics[scale=0.275]{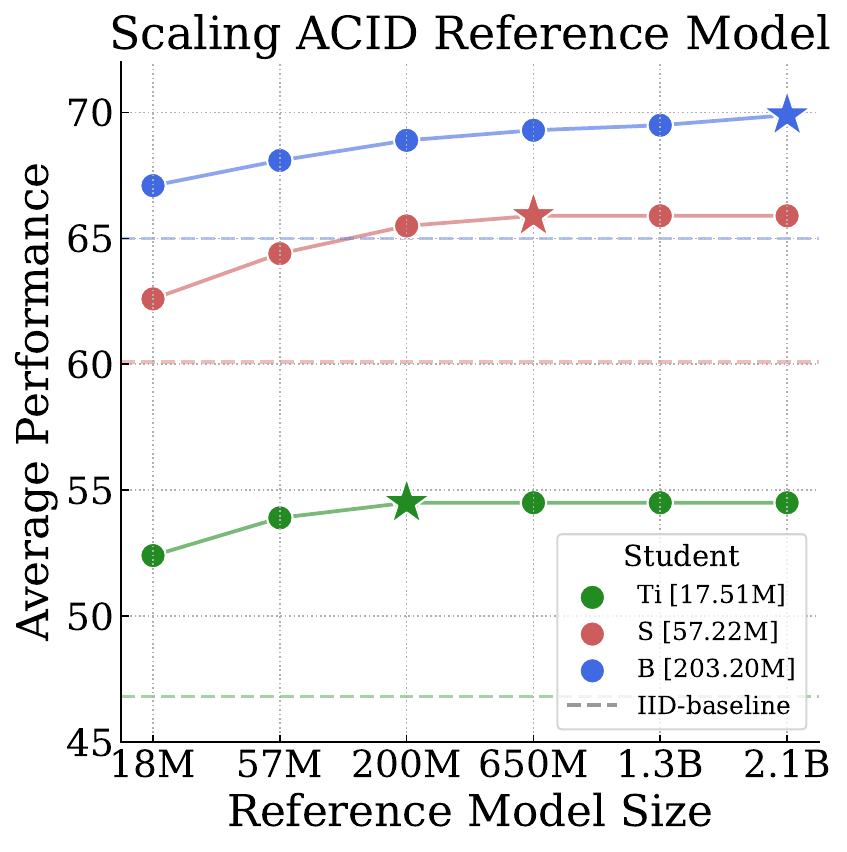}
    \label{fig:ref-model-scaling}
    ~
    \includegraphics[scale=0.275]{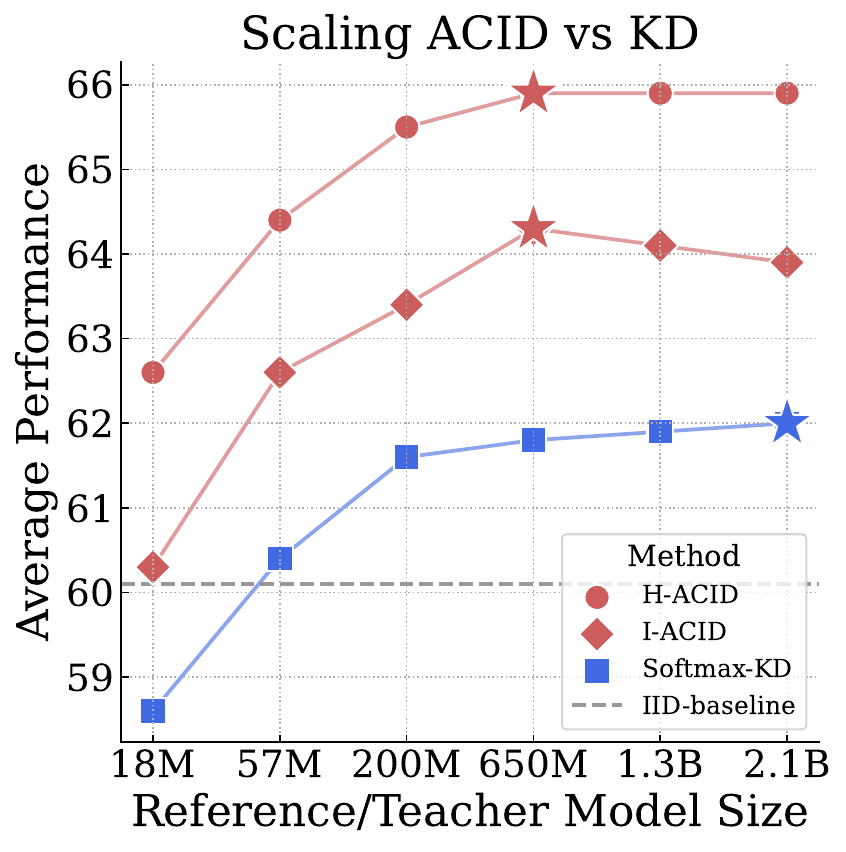}
    \label{fig:selectdist-vs-kd-scaling}
\vspace{-2em}
\caption{\textbf{Scaling behaviour of \ourmethod.} \textit{(left)} We scale up the reference model used for training each student (Ti, S and B) with \textcolor{Maroon}{\ourmethodhard}---there is an optimal scaling relationship (best reference for each student marked with $\star$) between student and reference sizes. \textit{(right)} Our \textcolor{Maroon}{\ourmethodhard} and \textcolor{Maroon}{\ourmethodeasy} comprehensively outperform \textcolor{RoyalBlue}{\textit{Softmax-KD}} across all teacher scales. Importantly, our \textcolor{Maroon}{\ourmethod}s outperform the \textcolor{Gray}{IID} baseline even for tiny reference models, whereas \textcolor{RoyalBlue}{\textit{Softmax-KD}} struggles to improve over IID with smaller teachers. \vspace{-1em}
}
\label{fig1}
\end{figure}

\vspace{0.5em}\noindent\textbf{Training Configurations.}
Unless otherwise specified, we train for $3$ billion total samples seen, with a batch-size of ${b}{=}{32,678}$ with the sigmoid contrastive loss (\cref{eq:sig-con}). The image-encoder takes images resized to $({256}{\times}{256})$ without additional augmentations.
The text-encoder uses a sentencepiece tokenizer~\citep{kudo2018sentencepiece} trained on English-C4~\citep{raffel2020exploring}, with a vocabulary size of $32,000$. 
We truncate all text captions to the first $64$ tokens. For most experiments, we use an rsqrt learning rate scheduler~\citep{zhai2022scaling}, with a peak learning-rate of $0.001$, and linear-warmup and linear-cooldown applied for $10\%$ of total steps.
By default, we use a filtering ratio of ${f}{=}{0.8}$ when using \ourmethod sampling, leading to a super-batch-size of ${B}{=}{163,840}$. We sweep over ${\lambda}{=}{\{{0.5}{,}{1.0}{,}{2.0}\}}$ for finding the optimal loss-weight for the \textit{Softmax-KD} loss (\cref{eq:smax-distillation-loss}).
For more details, refer to \cref{appx-sec:training}.

\subsection{\textbf{ACID} is an effective distillation method}
\label{sec:exp-acid-beats-distillation}
\subsubsection{Scaling behaviour}
To study the efficacy of \ourmethod as an effective distillation method, we first conduct a scaling study as the reference/teacher model size is increased. We use Hard-\ourmethod as our sampling scheme, and start with three fixed student models, Ti, S and B. We train each student by sweeping over (Ti, S, B, L, H and g) reference model sizes. Each reference model is trained on the WebLI-curated++ dataset for 2B samples seen, to ensure that the only difference across the experimental sweep is the size of the reference.~\cref{fig1} {(left)} showcases the scaling behaviour of each of the trained students, as the reference model is scaled up. We observe that across all student and reference models, our \ourmethod method always outperforms the IID-baseline (dotted lines). Moreover, we note that the best reference-student combination (highlighted with $\star$) changes as we scale up the student sizes---the B reference is best for the Ti student, L reference for S student, and g reference for B student. This suggests an optimal \textit{reference-student capacity ratio}---we can continue scaling up the reference model for \ourmethod sampling until we hit this capacity ratio, beyond which performance saturates. 

In~\cref{fig1} (right), we compare the scaling behaviour of our \ourmethod variants (both I- and H-) with the \textit{Softmax-KD} baseline, using an S student model. We note that across all reference/teacher scales, our \ourmethod methods are more effective at distilling the knowledge into the smaller S student. Moreover, both versions of our method outperform the IID baseline, even when using a smaller Ti reference model. Contrarily, \textit{Softmax-KD} only benefits when using much larger teacher models---this further demonstrates the scalability and flexibility of our \ourmethod distillation. Since \ourmethodhard demonstrates better scaling than \ourmethodeasy, we use that as our default in all further sections, and refer to it as our canonical \ourmethod (dropping the H- for better readability).

\begin{figure*}[h!]
    \centering
    \includegraphics[width=\textwidth]{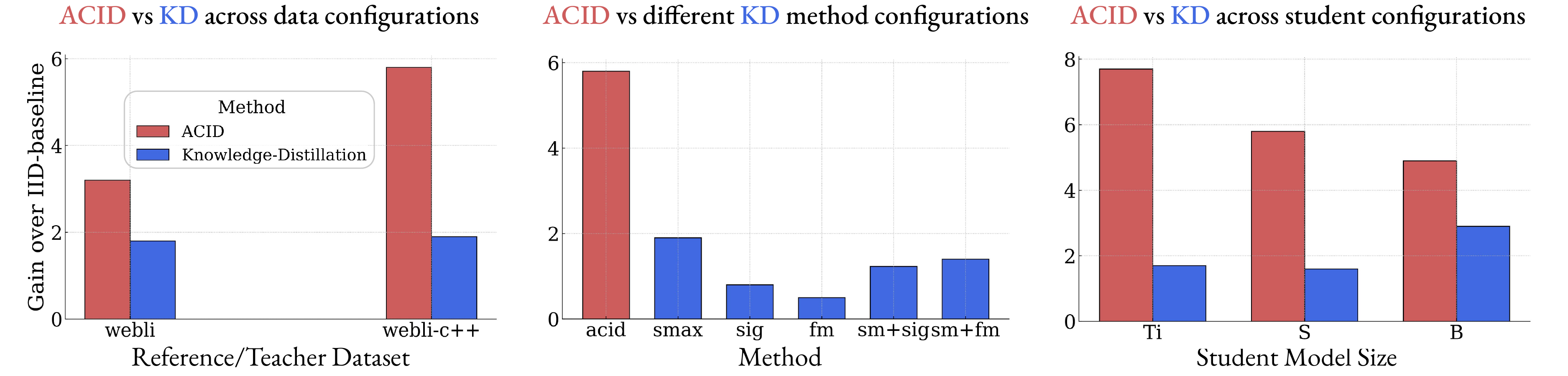}
    \vspace{-20pt}
    \caption{\textbf{\ourmethod significantly outperforms KD.} \textit{(left)} We vary the training dataset of the reference/teacher model, and use \textit{the same pretrained model} as the reference for \textit{ACID} and teacher for KD---across all configurations, we note strong gains for \ourmethod. \textit{(center)} Across different distillation objectives and a full hyperparameter sweep for optimal KD conditions, \ourmethod is still the best performing method by large margins. \textit{(right)} \ourmethod further outperforms KD across three different student sizes. \vspace{-1em}}
    \label{fig2}
\end{figure*}

\begin{figure}[htbp]
\centering
\includegraphics[width=\linewidth]{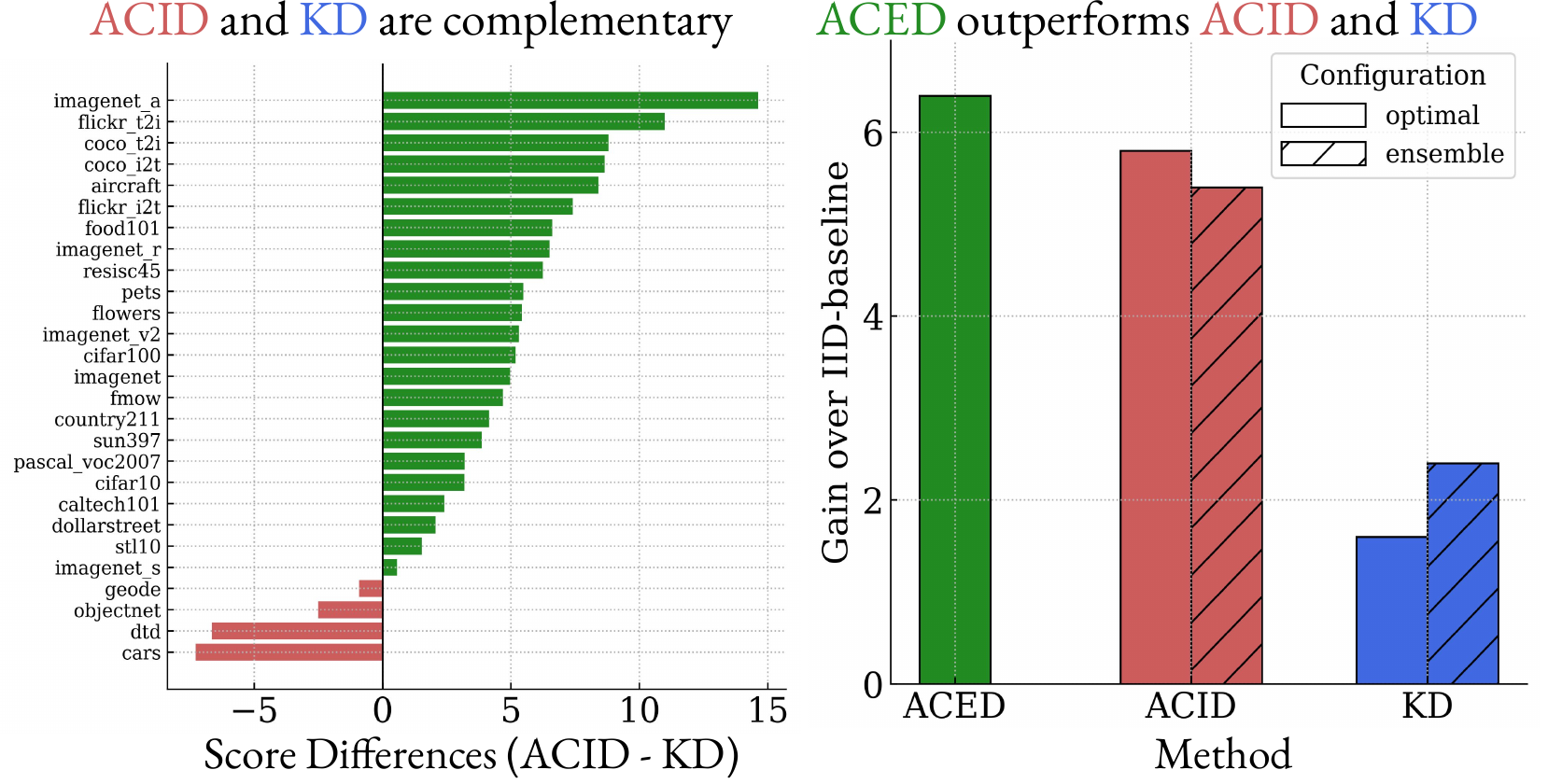}
\vspace{-15pt}
\caption{
\textbf{ACED for improved distillation.} 
\textit{(left)} Despite \ourmethod outperforming KD across most benchmarks, it still suffers on 4 out of 27 evals 
(potentially due to filtering out data). This motivates that combining \ourmethod and \textit{KD} would enable a stronger, more robust model. 
\textit{(right)} Our combined \ourmethodwithkd indeed outperforms both \ourmethod and \textit{KD}, even when using an ensemble of teacher/reference models for \ourmethod and \textit{KD}, showcasing its generality.\vspace{-1em}
}
\label{fig:selectdist2-motivation-and-results}
\end{figure}

\subsubsection{ACID outperforms standard distillation}
\label{sec:selectdist-outperform}
Having demonstrated the favourable scaling behaviour of \ourmethod vs. \textit{Softmax-KD} using a single teacher/reference model dataset, we next demonstrate that \ourmethod outperforms explicit distillation across different teacher/reference pretraining datasets, objective functions, and student model sizes.

\noindent\textbf{Reference/Teacher Training Dataset.} In~\cref{fig2} (left), we sweep over two different pretraining datasets for the references/teachers. We train an L-sized teacher/reference for 2B samples seen on WebLI-curated++ and WebLI. Using these models as teacher/reference, we train S students with \ourmethod, that strongly outperform \textit{Softmax-KD} for both datasets.

\noindent\textbf{Different Distillation Objectives.} Prior works have explored several different objectives for multimodal distillation, beyond standard \textit{Softmax-KD}. Here, we compare our \ourmethod method to some of these, including a \textit{Sigmoid-KD} loss~\citep{wu2023tinyclip} and a \textit{Feature-Matching KD} loss~\citep{yang2023clip} (see \cref{knowledge-distillation-appendix} for more details). Further, the SoTA multimodal distillation method, CLIP-KD~\citep{yang2023clip}, advocates combining these losses for best performance. We therefore also compare against two combination methods---\textit{Softmax+Sigmoid} and \textit{Softmax+Feature-Matching}. In~\cref{fig2} (center), we show that \ourmethod, without any additional complexity, still comprehensively outperforms all of the other distillation objectives.

\noindent\textbf{Different Student Sizes.} Finally, we also sweep across student sizes---Ti, S, and B. From~\cref{fig2} (right), we again observe that our \ourmethod substantially improves over \textit{Softmax-KD}. Interestingly, we note that our \ourmethod method is more effective for smaller students (Ti, S) than the larger B student, whereas this is the opposite for the \textit{Softmax-KD} baseline. 

\subsection{\textbf{{ACED}}: ACID and KD are complementary}

\noindent\textbf{Combining ACID and Softmax-Distillation---Why?} Theoretically in~\cref{sec:selectdist}, we show \ourmethod is in fact a form of implicit distillation, yet the exact form of this objective is different from traditional distillation. As a result, here we ask if this form of distillation (although stronger than traditional KD) is in fact complementary to standard distillation. 
This line of inquiry is further supported by an empirical finding shown in \cref{fig:selectdist2-motivation-and-results} \textit{(left)}---while \ourmethod outperforms \textit{Softmax-KD} by more than $5\%$ on tasks like COCO and Flickr retrieval, it underperforms \textit{Softmax-KD} on more finegrained evaluations like Cars and DTD. This suggests that despite the implicit distillation performed by \ourmethod, having an explicit distillation objective should further provide wider benefits.

\noindent\textbf{\ourmethodwithkd---How to combine?} We now discuss strategies for combining \ourmethod and \textit{Softmax-KD}. The simple strategy, \textit{ACIDistill}, samples a training batch using \ourmethod and applies both the contrastive and softmax-distillation loss on that batch. The alternative, 
\textit{IIDistill}, samples two batches independently, one with \ourmethod sampling and the other IID sampled, and applies the distillation loss on the IID batch while the contrastive loss is applied on the \ourmethod batch.
We study the scaling behaviour of both strategies by training ViT-S students with WebLI-L teachers and WebLI-curated++-references, for 3B, 6.5B and 13B samples seen. We observe \textit{ACIDistill} showcases better performance across all compute budget scales (see \cref{sec-acidistill}). 
Hence, going forward, we use \textit{ACIDistill} as the default strategy for combining \ourmethod and \textit{Softmax-KD}, and refer to that as our main \ourmethodwithkd method.

\noindent\textbf{How well does ACED perform?} We now compare our optimal \ourmethodwithkd method from before with the \ourmethod and \textit{Softmax-KD} methods applied independently. First, we find that our \ourmethodwithkd indeed outperforms both the independent methods, demonstrating that we are effectively able to leverage both the reference and teacher models.
As an additional ablation, we also conduct a comparison with an ensemble version of \ourmethod and \textit{Softmax-KD}, where we use both the WebLI-L and WebLI-curated++-L models as a two-teacher ensemble for \textit{Softmax-KD} and a two-reference ensemble for \ourmethod. We find that \ourmethodwithkd even outperforms these ensemble methods, suggesting that the benefits of our \ourmethodwithkd are not solely due to using multiple teacher and reference models, but rather due to 
optimally combining the two frameworks.

\begin{table*}[h!]
    \centering
    \resizebox{0.9\textwidth}{!}{
        \begin{tabular}{l|C{1.2cm}|c|cccccc|C{1.2cm}}
            \toprule[1.5pt]
            \multirow{2}{*}{\textbf{Method}} 
            & \multirow{2}{1.2cm}{\centering\textbf{Samples Seen}}
            & \multirow{2}{1.2cm}{\centering\textbf{Infer. GFlops}}
            & \multicolumn{4}{c}{\textbf{Zero-shot Classification}}
            & \multicolumn{2}{c}{\textbf{Retrieval}}
            & \multirow{2}{1.4cm}{\centering\textbf{Avg. Perf. (27 evals)}}
            \\
            \cmidrule(lr){4-7}
            \cmidrule(lr){8-9}
             & & &
             \imagenetval{} & IN-shift & Object-Centric & Scene-Centric & COCO & Flickr30k \\

             \midrule[1.25pt]

             \textcolor{gray}{DatologyAI-cls-S/32} & \textcolor{gray}{2.0B} & \textcolor{gray}{2.83}  & \textcolor{gray}{52.7}
              & \textcolor{gray}{36.6} & \textcolor{gray}{68.3} & \textcolor{gray}{47.0} & \textcolor{gray}{30.2} & \textcolor{gray}{48.6} & \textcolor{gray}{50.5} \\

             \textcolor{gray}{DatologyAI-ret-S/32} & \textcolor{gray}{2.0B} & \textcolor{gray}{2.83} & \textcolor{gray}{45.6}
              & \textcolor{gray}{35.9} & \textcolor{gray}{61.9} & \textcolor{gray}{44.9} & \textcolor{gray}{41.5} & \textcolor{gray}{64.0} & \textcolor{gray}{49.3} \\



             TinyCLIP-RN30M & 15.2B** & 6.93 & 59.1 & 43.0 & 70.2  & 52.7  & 43.3 & 71.2 & 56.6 \\


             TinyCLIP-45M/32 & 15.8B** & 3.70 & 62.7 & 48.3 & 74.8  & 56.6  & 45.4 & 72.1 &  60.4 \\



             TinyCLIP-63M/32 & 15.8B** & 5.65 & 64.5 & 50.4 & 76.4  & 58.3  & 47.7 & 75.5 & 62.1 \\


             MobileCLIP-S0 & 13B* & 3.70 &
             67.8 & 55.2 & 77.0 &  57.3 & 49.6 & 76.7 & 63.6 \\


             \rowcolor{DnCBG}{\textbf{{ACED}-F0}} & 13B & \textbf{3.30} & \textbf{68.5} & \textbf{56.1} & \textbf{77.9}  & \textbf{59.4}
             & \textbf{51.0} & \textbf{79.5} & \textbf{64.0} \\ 



             \midrule





             
           \textcolor{gray}{{DatologyAI-cls-B/32}} & \textcolor{gray}{5.1B} & \textcolor{gray}{7.39}  & \textcolor{gray}{63.2}
              & \textcolor{gray}{47.1} & \textcolor{gray}{75.4} & \textcolor{gray}{52.2} & \textcolor{gray}{38.5} & \textcolor{gray}{60.8}  & \textcolor{gray}{58.5} \\

             \textcolor{gray}{DatologyAI-ret-B/32} &  \textcolor{gray}{5.1B} & \textcolor{gray}{7.39}  & \textcolor{gray}{55.8}
              & \textcolor{gray}{45.9} & \textcolor{gray}{69.6} & \textcolor{gray}{53.5} & \textcolor{gray}{49.6} & \textcolor{gray}{72.6}  & \textcolor{gray}{57.3} \\

             CLIP-KD-RN50 & 0.5B & 9.09 & 54.9
              & 41.6 & 61.8 & 50.0 & 43.5 & 71.4 & 52.2 \\
              
             OpenAI-RN50 & 13B & 9.09 & 
             59.8 & 44.6 & 65.2 & 50.9 & 38.7 & 68.6 & 53.6 \\


             OpenAI-CLIP-B/32 & 13B & 7.39 & 63.3
              & 50.3 & 72.6 & 55.2 & 40.3 & 68.9 & 58.6 \\


             LAION-CLIP-B/32 & 34B & 7.39 & 66.6
              & 52.4 & 78.4 & 59.5 & 47.7 & 75.5 & 63.7 \\


             DataComp-CLIP-B/32 & 13B & 7.39 & 69.2
              & 56.1 & 80.0 & 59.3 & 45.4 & 70.1 & 64.6 \\




             MetaCLIP-CLIP-B/32 & 13B & 7.39 & 67.7
              & 55.1 & 77.9 & 59.2 & 46.7 & 73.0 & 63.9 \\

             CLIP-CID-B/32 & 7.2B & 7.39 & 62.7
              & 50.5 & (-) & (-) & (-) & (-) & (-) \\
              

            TinyCLIP-39M/16 & 20B** & 9.48  & 63.5 & 50.6 & 71.6  & 56.7  & 46.9 & 75.6 & 59.5  \\


             MobileCLIP-S1 & 13B* & 7.64 &
             72.6 & 63.3 & 80.4  & 61.6  & 53.0 & 80.0 & 67.9 \\


             \rowcolor{DnCBG}{\textbf{ACED-F1}} & 13B & \textbf{7.14} & \textbf{74.9} & \textbf{67.3} & \textbf{81.8}  & \textbf{64.0}  & \textbf{55.6} & \textbf{84.7} & \textbf{69.7} \\

             \midrule


             OpenAI-RN101 & 13B & 12.75 & 
             62.3 & 49.7 & 68.4 & 53.7 & 40.3 & 68.6 & 56.5 \\
             

             MobileCLIP-S2 & 13B* & 10.81 & 74.4 & 68.1 & 81.8  & 63.6  & 54.4 & 81.8 & 69.8 \\


          \rowcolor{DnCBG}{\textbf{ACED-F2}} & 13B & \textbf{10.29} & \textbf{76.9} & \textbf{70.7} & \textbf{82.3} & \textbf{64.6}  & \textbf{58.3} & \textbf{85.3} & \textbf{70.9} \\

             

             %
             %
             %

             %
             %
             %

             %
             %
             %
             



            \bottomrule[1.5pt]
        \end{tabular}
        }
    \vspace{-5pt}
    \caption{\textbf{\ourmethodwithkd outperforms all prior state-of-the-art methods.} We showcase results for our method at three different model inference-GFlop scales. Across all three model scales, the performance of our \ourmethodwithkd models improves on the prior SoTA across the 27 StableEval evaluation datasets, while using fewer inference FLOPs. For better interpretability, we also break down the evaluations into individual benchmarks (\textit{e.g.}, ImageNet, COCO) and groupings (\textit{e.g.}, ImageNet distribution shifts, other object categorization, and scene classification---for details, see \cref{appx-sec:evaluation}). $^*$MobileCLIP samples seen include two captions for each image, effectively doubling the total unique pairs.$^{**}$TinyCLIP models are not trained from scratch, but use a complex weight inheritance strategy from pretrained models.}
    \label{tab:full_eval}
    \vspace{-5pt}
\end{table*}

\begin{table}[h!]
    \centering
    \vspace{-7pt}
    \resizebox{0.97\linewidth}{!}{
        \begin{tabular}{l|C{1.2cm}|c|c|cc}
            \toprule[1.5pt]
            \multirow{2}{*}{\textbf{Method}} 
            & \multirow{2}{1.2cm}{\centering\textbf{Samples Seen}}
            & \multirow{2}{1.2cm}{\centering\textbf{Image GFlops}}
            & {\centering\textbf{Captioning}}
            & \multicolumn{2}{c}{\textbf{VQA}}
            \\
            \cmidrule(lr){4-6}
             & & & Flickr30k &
            VQAv2 & GQA \\

             \midrule[1.25pt]






             SigLIP-B/16 & 40B & 23.45 & 53.4 & 64.5 & 54.9 \\

             SiLC-B/16 & 20B & 23.45 & 49.2 &  65.7 & 54.1  \\


             \rowcolor{DnCBG}{\textbf{ACED (B/16)}} & 13B & \textbf{23.19}  & \textbf{55.5} & \textbf{66.6} & \textbf{55.4} \\
             
            \bottomrule[1.5pt]
        \end{tabular}
        }
    \vspace{-5pt}
    \caption{\textbf{LiT-Decoder Evaluations.} Our ACED vision-encoders also improve performance of multimodal-decders on captioning and VQA tasks, when compared to strong SigLIP and SiLC baselines.}
    \vspace{-5pt}
    \label{tab:it_decoder_eval}
    \vspace{-5pt}
\end{table}

\subsection{Comparison to Prior Art}

We now pretrain \ourmethodwithkd models at large compute budgets, across three FLOP-scales, and compare with SoTA inference-efficient two-tower VLMs, including MobileCLIP~\citep{vasu2024mobileclip}, TinyCLIP~\citep{wu2023tinyclip}, CLIP-KD~\citep{yang2023clip}, CLIP-CID~\citep{yang2024clip} and \textcolor{gray}{proprietary DatologyAI-CLIP}~\citep{datologyaimultimodaldatacuration} (see \cref{expansion-on-model-details}). 
We train our \ourmethodwithkd-F0, \ourmethodwithkd-F1 and \ourmethodwithkd-F2 models on DataComp-1B for 13B samples seen.
From~\cref{tab:full_eval}, we observe that our \ourmethodwithkd models are the most FLOP-efficient and highly
performant---\ourmethodwithkd-F0 outperforms MobileCLIP-S0 by $0.4$\% and TinyCLIP-63M/32 by $1.9$\%, on average, while being $10.81$\% and $41.5$\% more FLOP-efficient respectively; our \ourmethodwithkd-F1 outperforms MobileCLIP-S1 by $1.8$\%, and TinyCLIP-39M/16 by $10.2$\%, on average, while being $6.5$\% and $24.6$\% more efficient respectively; \ourmethodwithkd-F2 outperforms MobileCLIP-S2 by $1.1$\% on average, while being $4.8$\% more FLOP-efficient.
Notably, on the most widely-used evaluations like ImageNet, COCO and Flickr, our method surpasses the previous SoTA by large margins---\ourmethodwithkd-F0 outperforms MobileCLIP-S0 by $0.7$\% on ImageNet, $1.5$\% on COCO, and $2.8$\% on Flickr; \ourmethodwithkd-F1 outperforms MobileCLIP-S1 by $2.3$\% on ImageNet, $2.6$\% on COCO, and $4.7$\% on Flickr, while \ourmethodwithkd-F2 outperforms MobileCLIP-S2 by $2.5$\% on ImageNet, $3.9$\% on COCO, and $3.5$\% on Flickr. 
In fact, our \ourmethodwithkd-F1 model even outperforms MobileCLIP-S2 on ImageNet, while having $34$\% lesser GFlops (see~\cref{fig:aced_teaser}).
This further validates the scalability of our \ourmethodwithkd, especially given our models \textit{do not} use any bespoke architectures or complex augmentations.

\subsection{ACED yields better encoders for other tasks}
We next evaluate the benefits of \ourmethodwithkd specifically for training an auto-regressive text-decoder with a frozen image-encoder, in the LiT-Decoder setting~\citep{beyer2023study}.
We evaluate the trained models on Flickr30k \citep{plummer2015flickr30k} captioning (using CIDEr \citep{vedantam2015cider}) and visual-question-answering (using accuracy).
Since prior works on these benchmarks use larger foundation models (\textit{e.g.}, SigLIP~\citep{zhai2023sigmoid} and SiLC~\citep{naeem2023silc}), we also train a larger \ourmethodwithkd (B/16) model for 13B samples seen 
\cref{tab:it_decoder_eval} demonstrates that our \ourmethodwithkd model outperforms both strong baselines across both tasks---particularly, our model outperforms competitors that have similar image-GFlops but are trained for a \textit{significantly higher number of samples seen} (up to $\sim$3x). 
This further highlights the impact of \ourmethodwithkd particularly for distilling knowledge in the image-encoders.



 



\section{Conclusion}
\label{sec:disc}
In this work we showed that 
active data curation implicitly implements a novel form of distillation, which combines knowledge from both a reference model and the data itself. 
With this insight, we developed \ourmethod, a powerful method for distilling large multimodal encoders into much more efficient ones 
via online joint-example selection~\citep{evans2024data}. 
\ourmethod 
strictly outperforms traditional forms of knowledge distillation in training contrastive VLMs.
Given
that \ourmethod implicitly optimizes a different objective than traditional softmax-based KD, we further demonstrated these two objectives to be complementary, 
arriving at our final method, \ourmethodwithkd, which combines the benefits of each. Using \ourmethodwithkd we distilled models that set a new state-of-the-art for FLOP-efficient zero-shot classification and image-text retrieval.  



\vspace{0.3cm}
\noindent\textbf{Acknowledgements.} The authors would like to thank (in alphabetic order of first name) Alexander Kolesnikov, André Susano Pinto, Andrew Zisserman, Diego Martin Arroyo, Karsten Roth, Lucas Beyer, Marco Fornoni, Tianshi Cao, and Xiaohua Zhai for helpful comments, feedback and support throughout the project.
{
    \small
    \bibliographystyle{ieeenat_fullname}
    \bibliography{main}
}

\appendix
\onecolumn
\clearpage
\setcounter{page}{1}

\appendix


\section{Evaluation Protocol Details}
\label{appx-sec:evaluation}



In the main text in~\cref{sec:impdetails}, we described the motivation and methodology for choosing our \textit{StableEval} set of 27 evaluations. We also categorized our main results in~\cref{tab:full_eval} into IN-shift, Object-Centric, and Scene-Centric, under the zero-shot classification section. We now provide additional details for these sections.\vspace{0.3cm}

\noindent\textbf{\textit{StableEval} Protocol.} For rigorously defining our final evaluation suite, we first selected $34$ candidate evaluation datasets popularly used for evaluating standard image-text contrastive pretraining~\citep{radford2021learning,gadre2024datacomp,lavoie2024modeling} and adaptation~\citep{zhou2022learning,gao2024clip,zhang2022tip,roth2023waffling,udandarao2023sus} methods. These datasets ranged from standard natural image-classification, to fine-grained classification of birds, animals, and cars etc., to different domains of images like satellite imagery and street signs. The full set of $34$ candidate evaluations we started with are: FGVC-Aircrafts \citep{maji13fine-grained}, Oxford Flowers-102 \citep{nilsback2008automated}, Oxford-IIIT Pets \citep{oxfordpets}, Stanford Cars \citep{krause20133d}, Food-101 \citep{bossard2014food}, Caltech-101 \citep{caltech101}, CIFAR-10 \citep{krizhevsky2009learning}, CIFAR-100 \citep{krizhevsky2009learning}, Pascal VOC 2007 \citep{pascal-voc-2007}, EuroSAT \citep{eurosat}, RESISC45 \citep{resisc45}, STL-10 \citep{coates2011analysis}, SUN-397 \citep{xiao2010sun}, Dollar Street \citep{dollarstreet}, GeoDE \citep{ramaswamy2024geode}, Country211 \citep{radford2021learning}, FMoW \citep{christie2018functional}, DTD \citep{cimpoi14describing}, iWildCam \citep{beery2021iwildcam}, PatchCamelyon \citep{veeling2018rotation}, CLEVR Counts \citep{johnson2017clevr}, CLEVR Distance \citep{johnson2017clevr}, KITTI Distance \citep{geiger2012we}, ImageNet-V2 \citep{imagenetv2}, ImageNet-A \citep{hendrycks2021natural}, ImageNet-R \citep{hendrycks2021many}, ObjectNet \cite{objectnet}, ImageNet-Val \citep{deng2009imagenet}, ImageNet-Sketch \citep{wang2019learning}, Rendered SST2 \citep{radford2021learning}, Flickr30k (I2T and T2I) \citep{plummer2015flickr30k}, MSCOCO ((I2T and T2I)) \citep{lin2014microsoft}. 

We then trained several variants of standard SigLIP and CLIP models with a ViT-S/32 image-encoder and a BERT-small text-encoder, to quantify the amount of variance present for each evaluation dataset, solely due to the random seed (\textit{i.e.}, different initialization of model weights). Specifically, we first trained $5$ IID-SigLIP models on both DataComp-1B and WebLI-1B for 3B examples seen (\textit{i.e.}, with randomly sampling batches of data at each step) by only changing the random seed. Note that we ensured that the exact samples seen per step in the training process was fixed---that is, the only randomness across the $5$ different seed runs was the model initialization. We also trained an IID-CLIP model for 5 seeds to add variation on the training objective to the set of models. We then get the average standard deviation of each evaluation dataset by first averaging over the $5$ different random seeds per method (\textit{i.e.}, DataComp-IID-SigLIP, DataComp-IID-CLIP, WebLI-IID-SigLIP), and then averaging over the $3$ different combinations of methods. This average standard deviation is taken to be the variability of each evaluation, which is shown in~\cref{fig:stable-eval}. We also tested this variability across other settings by changing the patch-size of the image-encoder (from S/32 to S/16) and increasing the model size (from S/32 to B/32), and found the variability (standard deviation) per evaluation dataset to be consistent. 

Equipped with these standard deviations per evaluation dataset, we then aim to prune out the set of highly unstable evaluations from the full set of $34$ evaluations by taking inspiration from the continuous inverse-variance weighting (IVW) method~\citep{hartung2011statistical}. We start with the lowest-variance evaluation (Country211 with $0.15$\% standard deviation), and progressively add evaluations in increasing order of their computed standard deviations, each time computing the variability of the average over the current set of evaluations. For a set of $N$ evaluations, the variability of the average is computed as ${std}{(E_1 ... E_N)}{=}{\sqrt{\frac{1}{N^2}\sum_i{var(E_i)}}}$. At each step, we compare the variability of the average with the variability of the most reliable evaluation (\textit{i.e.}, Country211 with $0.15$\% standard deviation), and prune out all evaluations beyond the critical point where the variability of the average becomes larger than the Country211 variability. This leaves us with a set of $27$ evaluations that are both diverse as well as stable across different random seeds. The $7$ evaluation datasets that were pruned out of the final set are: EuroSAT, CLEVR Counts, GTSRB, iWildCam, SVHN, KITTI Distance, CLEVR Distance, PatchCamelyon, and Rendered-SST2. \vspace{0.3cm}

\noindent\textbf{Categorization of Datasets.} Having identified our stable set of evaluation datasets, we next categorize them into different brackets for easier parsing of the different capabilities of the models in~\cref{tab:full_eval}. In~\cref{tab:evaluation-sets}, we showcase the breakdown of the different categories represented in~\cref{tab:full_eval} for all $27$ evaluations. We categorize them into \textit{object-centric} datasets like FGVC-Aircrafts or Stanford Cars, \textit{scene-centric} datasets like SUN-397 or RESISC45, \textit{Imagenet-based natural distribution shifts} like ImageNet-V2 or ObjectNet, and other \textit{miscellaneous} evaluations like DTD or Country211. Finally, we also evaluate our models on image-text retrieval datasets like COCO and Flickr, both using text-to-image retrieval and image-to-text retrieval, as separate evaluation metrics.

\begin{table*}
\centering
\caption{\textbf{Final \textit{StableEval} Set of $27$ evaluations.}}
\setlength\tabcolsep{4.5pt}
    \renewcommand{\arraystretch}{1.1}
        \rowcolors{2}{white}{}
\resizebox{0.8\textwidth}{!}{
\centering
\begin{tabular}{lll|rr}
\toprule[1.5pt]
\textbf{Category} & \textbf{Dataset} & \textbf{Task} & \textbf{Test set size} & \textbf{Number of classes} \\ \midrule[0.7pt]
\cellcolor{white} & FGVC-Aircrafts \citep{maji13fine-grained} & Aircraft recognition & 3,333 & 100 \\
\cellcolor{white} & Oxford Flowers-102 \citep{nilsback2008automated} & Flower recognition & 6,149 & 102 \\
\cellcolor{white} & Oxford-IIIT Pets \citep{oxfordpets} & Pet classification & 3,669 & 37 \\
\cellcolor{white} & Stanford Cars \citep{krause20133d} & Vehicle recognition & 8,041 & 196 \\
\cellcolor{white} & Food-101 \citep{bossard2014food} & Food recognition & 25,250 & 101 \\
\cellcolor{white}  & Caltech-101 \citep{caltech101} & Object recognition & 6,085 & 102 \\
\cellcolor{white} & CIFAR-10 \citep{krizhevsky2009learning} & Visual recognition & 10,000 & 10 \\
\cellcolor{white} & CIFAR-100 \citep{krizhevsky2009learning} & Visual recognition & 10,000 & 100 \\
\cellcolor{white} & Pascal VOC 2007 \citep{pascal-voc-2007} & Object recognition & 14,976 & 20 \\
\cellcolor{white} \multirow{-10}{*}{Object-Centric} & STL-10 \citep{coates2011analysis} & Visual recognition & 8,000 & 10 \\
\cmidrule{1-5}

\cellcolor{white} & SUN-397 \citep{xiao2010sun} & Scene recognition & 108,754 & 397  \\
\cellcolor{white} & GeoDE \citep{ramaswamy2024geode} & Object/scene recognition & 12,488 & 40 \\
\cellcolor{white} & RESISC45 \citep{resisc45} & Satellite imagery recognition & 6,300 & 45 \\
\cellcolor{white} \multirow{-4}{*}{Scene-Centric}& FMoW \citep{christie2018functional} & Satellite imagery recognition & 22,108 & 62 \\
\cmidrule{1-5}

\cellcolor{white} & ImageNet-V2 \citep{imagenetv2} & Visual recognition & 10,000 & 1,000 \\
\cellcolor{white} & ImageNet-A \citep{hendrycks2021natural} & Visual recognition & 7,500 & 200 \\
\cellcolor{white} & ImageNet-R \citep{hendrycks2021many} & Visual recognition & 30,000 & 200  \\
\cellcolor{white} \multirow{-4}{*}{Distribution-shifts}& ObjectNet \cite{objectnet} & Visual recognition & 18,574 & 113 \\
\cmidrule{1-5}

\cellcolor{white} & ImageNet-Val \citep{deng2009imagenet} & Visual recognition & 50,000 & 1,000 \\
\cellcolor{white} & ImageNet-Sketch \citep{wang2019learning} & Visual recognition & 50,889 & 1,000 \\
\cellcolor{white} & DTD \citep{cimpoi14describing} & Texture classification & 1,880 & 47 \\
\cellcolor{white} & DollarStreet \citep{dollarstreet} & Object recognition & 3,503 & 58 \\
\cellcolor{white} \multirow{-5}{*}{Misc.} & Country211 \citep{radford2021learning} & Geolocation & 21,100 & 211 \\

\midrule
 \cellcolor{white} & Flickr30k (I2T, T2I) \citep{plummer2015flickr30k} & Image and text retrieval & 31,014 & N/A  \\
\cellcolor{white} \multirow{-2}{*}{Retrieval}& MSCOCO (I2T, T2I) \citep{lin2014microsoft} & Image and text retrieval & 5,000 & N/A \\
\bottomrule[1.5pt]
\end{tabular}} 
\label{tab:evaluation-sets}
\end{table*}

\newpage
\section{Image-text contrastive Objectives}
\label{contrastive-loss-appendix}

Here, we expand the full image-text pretraining objectives described in~\cref{sec:preliminaries}. The per-sample softmax image-text objective is primarily used for training CLIP~\citep{radford2021learning} models, while the per-sample sigmoid objective is primarily used in training SigLIP~\citep{zhai2023sigmoid} models:

\begin{align}
\label{eq:smax-con}
\mathcal{L}_{\text{softmax}} (x_i; \mathcal{B}) & = 
- \frac{1}{2} \left( \log p_{ii}^{\text{img} \rightarrow \text{txt}}  + \log p_{ii}^{\text{txt} \rightarrow \text{img}} \right) \\ 
\label{eq:sig-con}
\mathcal{L}_{\text{sigmoid}} (x_i; \mathcal{B}) & = 
- \left( \log p_{ii}^\text{sig} + \sum_{{j}{=}{1}{,}{j}{\neq}{i}}^{b} \log (1 - p_{ij}^\text{sig}) \right)
\end{align}

\newpage
\section{Proofs for Active Curation as Implicit Distillation}
\label{acid-derivation-appendix}

In this section, we provide derivations for our theoretical results in~\cref{sec:selectdist} showcasing the equivalence between active data curation and knowledge distillation. We first show the proof for the case where we use easy-reference scoring for data-curation, followed by the learnability-scoring case, and finally showcase a generalized version of the proof.\vspace{0.3cm}

\noindent\textbf{Setup.} Recollect from the main paper text in~\cref{sec:selectdist}, that we are given an image-text pretraining dataset $\mathcal{D}$. The simple training approach is to sample uniformly random batches of data $\mathcal{B}$ (of size $b$), from $\mathcal{D}$ at each step $t$, and minimize $\mathcal{L} \in \{ \mathcal{L}_{\text{softmax}}, \mathcal{L}_{\text{sigmoid}}\}$ (see~\cref{contrastive-loss-appendix} for full equations for the loss objectives). We call this baseline, minimizing $\hat{\mathcal{L}} = \frac{1}{b} \sum_{x_i \sim \mathcal{U}[\mathcal{D}]} \mathcal{L}(x_i; \mathcal{B})$ as the \textit{IID-baseline} ($\theta_{\text{IID}}$). 
Further, remember that in the \textit{active data curation} setup, we employ a smarter way to select batches, using a pretrained \textit{reference} model $\theta_{\text{ref}}$. At each step $t$, we select a sub-batch $\mathcal{B}$ (size $b$) from a much larger super-batch $\mathcal{S}$ (size $B$) according to an \textit{active selection distribution} $\mathcal{A}[\mathcal{S}]$.\vspace{0.3cm}

\noindent\textbf{\underline{A}ctive Data \underline{C}uration as \underline{I}mplicit \underline{D}istillation (ACID).} We now show formally that 
active curation can be cast as ``implicit distillation'' and should benefit from larger reference models. 
The model now minimizes 
$\hat{\mathcal{L}} = \frac{1}{b} \sum_{x_i \sim \mathcal{A}[\mathcal{S}]} \mathcal{L}(x_i; \mathcal{B})$, which in expectation is $ \mathcal{E} = \mathbb{E}[\hat{\mathcal{L}}] = \sum_{x\in\mathcal{D}} a(x) \mathcal{L}(x; \mathcal{B})$ given that super-batches $\mathcal{S}$ are sampled uniformly.
Recall that $\mathcal{L}(x; \mathcal{B}) = -\sum_{i=1}^b y_i(x) \log q_i(x)$
, where $y_i$ are the labels of the contrastive task and $q_i$ are the probabilities induced by the pairwise similarities of the student $\theta$. Let $p_i$ be the probabilities induced by the reference model $\theta_{\text{ref}}$. In the case of \textit{easy-reference scoring} and the softmax loss, $a(x) = \frac{1}{Z} \exp \sum_{i=1}^b y_i(x) \log p_i(x) = \frac{1}{Z} p_{i^*}(x)$ where $i^*$ is the index of the one-hot label $y(x)$. As such,
\begin{align}
    \mathcal{E}_\text{easy-ref} & = - \sum_{x \in \mathcal{D}} a(x) \sum_{i=1}^b y_i(x) \log q_i(x) \\ & = - \frac{1}{Z} \sum_{x \in \mathcal{D}} p_{i^*}(x) \sum_{i=1}^b y_i(x) \log q_i(x) \nonumber \\ 
    \label{eq:acid-proof}
    & = - \frac{1}{Z} \sum_{x \in \mathcal{D}} \sum_{i=1}^b p_{i^*}(x) y_i(x) \log q_i(x) \nonumber \\ & = \frac{1}{Z} \sum_{x \in \mathcal{D}} \text{KD}[ p(x) \cdot y(x) ; q(x) ] \nonumber
\end{align}
This demonstrates that by curating data according to
the reference model $\theta_{\text{ref}}$, we implicitly distill its knowledge via a novel data-driven objective, using a
combination of model predictions and real labels as targets. We next prove the equivalence of data curation and knowledge-distillation, when using learnability-based scoring for our active data curation.

\vspace{0.5em}\noindent\textbf{Learnability-based Data Curation is Hard Distillation.} When using learnability-based prioritization, the active selection distribution $\mathcal{A}$ factorizes as 
$a^\text{learn} = \frac{1}{Z} \exp(s^\text{learn}) = \frac{1}{Z} \exp[\mathcal{L}(\cdot | \theta) - \mathcal{L}(\cdot | \theta_\text{ref})] = a^\text{easy-ref} \cdot a^\text{hard-learn}$ where $a^\text{hard-learn} = \frac{1}{Z} \exp[\mathcal{L}(\cdot | \theta)]$ prioritizes examples with high loss according to the student.
Since easy-reference prioritization yields implicit distillation (\ourmethodeasy, \cref{eq:acid}), learnability prioritization yields:
\begin{align}    
    \mathcal{E}_\text{learn} & = \sum_{x \in \mathcal{D}}  a^\text{hard-learn}(x) \cdot a^\text{easy-ref}(x) \mathcal{L}(x; \mathcal{B}) \\ & = \frac{1}{Z} \sum_{x \in \mathcal{D}}  a^\text{hard-learn}(x) \text{KD}[ p(x) \cdot y(x) ; q(x) ] \nonumber
\end{align}
\noindent This demonstrates that learnability-based active curation is equivalent to implicit distillation on hard examples (``H-ACID'') according to the student model.

\noindent\textbf{ACID for general learning objectives.} In the general case (including sigmoid-contrastive learning, and combined image-to-text and text-to-image softmax contrastive learning), $y(x)$ contains a set of labels $y_i(x)$ such that $\sum_{i=1}^b y_i(x) = 1$. In this case
$ a(x) = \frac{1}{Z} \exp \sum_{i=1}^b y_i(x) \log p_i(x) 
 \leq \frac{1}{Z} \sum_{i=1}^b  y_i(x) p_i(x) = \frac{1}{Z} \hat{p}(x)$ due to the convexity of the exponential. In particular, 
 \begin{align}
      \mathcal{E}_\text{easy-ref} = - \sum_{x \in \mathcal{D}} a(x) \sum_{i=1}^b y_i(x) \log q_i(x) & \geq - \frac{1}{Z} \sum_{x \in \mathcal{D}} \hat{p}(x) \sum_{i=1}^b y_i(x) \log q_i(x) \\
      & \geq \frac{1}{Z} \sum_{x \in \mathcal{D}} \text{KD}[ \hat{p}(x) \cdot y(x) ; q(x) ]
 \end{align}
As such, learning from actively-curated data minimizes an upper bound on the KD objective described previously, for general learning objectives of the form  $\sum_{i=1}^b y_i(x) \log q_i(x)$, including the softmax- and sigmoid-contastive objectives we utilize in this work. 

\newpage
\section{Knowledge Distillation Objectives}
\label{knowledge-distillation-appendix}

In this section, we describe in detail all the knowledge-distillation methods we use to compare as baselines in our results in~\cref{sec:exp-acid-beats-distillation}. Given the student model $\theta$ and a pretrained teacher model $\theta_{\text{teacher}}$, we considered three main objectives for distilling the knowledge from the teacher $\theta_{\text{teacher}}$ into the student model $\theta$.\vspace{0.3cm}

\noindent\textbf{Softmax contrastive distillation.} Here, our aim is to distill the contrastive logit matrix from the teacher to the student. Formally, given a data-batch $B$, we extract teacher embeddings $\{({z}^{\text{img}}_{i,t},{z}^{\text{txt}}_{i,t})\}$ and student embeddings $\{({z}^{\text{img}}_{i,s},{z}^{\text{txt}}_{i,s})\}$. The teacher and student contrastive matrices, $\mathcal{T}_{b\times b}$ and $\mathcal{S}_{b\times b}$, contain the teacher and student image-text logits, respectively:
\begin{equation}
\label{eq:contrastive_matrix-appx}
\mathcal{T}_{i,j}={\alpha}_{{t}}{z}^{\text{img}}_{i,t}\cdot{z}^{\text{txt}}_{j,t}  ,   \mathcal{S}_{i,j}={\alpha}_{{s}}{z}^{\text{img}}_{i,s}\cdot{z}^{\text{txt}}_{j,s}
\end{equation}

Our softmax distillation objective takes the form of a cross-entropy loss between the teacher and student contrastive matrices, considering the texts as labels by applying a row-wise softmax on the contrastive matrices ($\mathcal{T}$, $\mathcal{S}$) and the images as labels by applying a column-wise softmax ($\mathcal{T}^T$, $\mathcal{S}^T$). 

\begin{equation}
\label{eq:smax-distillation-loss-appx}
\mathcal{L}_{\text{smax-dist}} = -\frac{1}{2b}\sum_{i=1}^{b}\left(\underbrace{\texttt{softmax}(\mathcal{T}_{i,\cdot})\log\texttt{softmax}(\mathcal{S}_{i,\cdot})}_{\text{image-to-text}}\right.
+
\left.\underbrace{\texttt{softmax}(\mathcal{T}_{i,\cdot}^{T})\log\texttt{softmax}(\mathcal{S}_{i,\cdot}^{T})}_{\text{text-to-image}}
\right)
\end{equation}
\vspace{0.3cm}

\noindent\textbf{Sigmoid contrastive distillation.} Similarly as above, here we distill the teacher contrastive matrix into the student matrix. However, differently from the softmax case, in this loss we use the full teacher and student image-text logits with the addition of the bias term:
\begin{equation}
\label{eq:contrastive_matrix-for-sigmoid}
\mathcal{T}_{i,j}={\alpha}_{{t}}{z}^{\text{img}}_{i,t}\cdot{z}^{\text{txt}}_{j,t}{+}{\beta_{{t}}}  ,   \mathcal{S}_{i,j}={\alpha}_{{s}}{z}^{\text{img}}_{i,s}\cdot{z}^{\text{txt}}_{j,s}{+}{\beta_{{s}}}
\end{equation}
Our sigmoid distillation objective then simply takes the form a binary cross-entropy objective between the teacher and the student logits (converted to probabilites using the sigmoid ($\sigma$) activation):
\begin{equation}
\label{eq:sigmoid-distillation-loss}
\mathcal{L}_{\text{sig-dist}} = -\frac{1}{b}\sum_{i=1}^{b}\Biggl( \sigma(\mathcal{T}_{i,\cdot})\log\sigma(\mathcal{S}_{i,\cdot}) + \sigma(-\mathcal{T}_{i,\cdot})\log\sigma(-\mathcal{S}_{i,\cdot}) \Biggl)
\end{equation}
\vspace{0.3cm}

\noindent\textbf{Feature-matching distillation.} We also explore a distillation loss that directly aligns the image and text embeddings of the student and teacher models directly, using a simple mean-squared error. Such a strategy has also been explored in prior SoTA CLIP distillation works~\citep{yang2023clip}, with great efficacy. If the student and teacher embedding dimensions are different, we project the student embedding to the teacher dimension using a learnable linear projection head ${P}_{\text{head}}$:
\begin{equation}
\begin{aligned}
\label{eq:feat-matching-distillation-loss}
{\hat{z}}^{\text{img}}_{i,s}={{z}}^{\text{img}}_{i,s}{P}_{\text{head}}, {\hat{z}}^{\text{txt}}_{i,s}={{z}}^{\text{txt}}_{i,s}{P}_{\text{head}}\\
\mathcal{L}_{\text{fm-dist}} = \frac{1}{2b}\sum_{i=1}^{b}\Biggl(\underbrace{\| {\hat{z}}^{\text{img}}_{i,s}-{{z}}^{\text{img}}_{i,t} \|_{2}^{2}}_{\text{image align}}{+}\underbrace{\| {\hat{z}}^{\text{txt}}_{i,s}-{{z}}^{\text{txt}}_{i,t} \|_{2}^{2}}_{\text{text align}}\Biggl)
\end{aligned}
\end{equation}
\vspace{0.3cm}

\noindent\textbf{Students with Knowledge Distillation.} For training student models with KD-objectives as specified above, we always use them in conjunction with the standard contrastive loss (either~\cref{eq:smax-con} or~\cref{eq:sig-con}): 

\begin{equation}
\label{eq:kd-only-baseline-loss}
\mathcal{L}_{\text{dist-only}} = \mathcal{L}_{\text{softmax/sigmoid}} +\lambda_{\text{smax}}\cdot\mathcal{L}_{\text{smax-dist}} + \lambda_{\text{sig}}\cdot\mathcal{L}_{\text{sig-dist}} + \lambda_{\text{fm}}\cdot\mathcal{L}_{\text{fm-dist}}
\end{equation}

This objective allows us to flexibly combine the different distillation objectives by varying the different loss-weights $\lambda_{\text{smax/sig/fm}}$. By default, we use only the softmax distillation objective with a loss-weight of $2.0$, however we perform sweeps over multiple configurations of loss-weights and loss-combinations in our experiments.\vspace{0.3cm}

\noindent\textbf{Ensemble Teachers.} The above distillation setup also easily enables using multiple teacher models in an ensemble for teaching the student. Such an ensemble teacher strategy has been explored in prior SoTA multimodal distillation works~\cite{vasu2024mobileclip}. For a teacher ensemble, the distillation objective simply averages the predicted logits from the different teachers. As an example, an ensemble-softmax-distillation objective would be as follows:

\begin{multline}
\label{eq:ensemble-teachers}
\mathcal{L}_{\text{ens-smax-dist}} = -\frac{1}{2bK}\sum_{k=1}^{K}\sum_{i=1}^{b}( \underbrace{\texttt{softmax}(\mathcal{T}_{i,\cdot}^{k})\log\texttt{softmax}(\mathcal{S}_{i,\cdot})}_{\text{image-to-text}}
{+}
\underbrace{\texttt{softmax}({\mathcal{T}_{i,\cdot}^{k}}^{T})\log\texttt{softmax}(\mathcal{S}_{i,\cdot}^{T})}_{\text{text-to-image}})
\end{multline}

\newpage
\section{Training Details}
\label{appx-sec:training}

Our default configuration follows that of SigLIP~\citep{zhai2023sigmoid}.
Unless otherwise specified, we train for $3$ billion total samples seen, with a batch-size of ${b}{=}{32,678}$ with the sigmoid contrastive loss (\cref{eq:sig-con}). The image-encoder takes images resized to $({256}{\times}{256})$ without any additional augmentations. By default for all our ablation experiments, we use a ViT-S/16 image encoder and a BERT-small text encoder. The image encoder uses global-average pooling (GAP) for the final embedding by default, however for some experiments we also use multi-head attention pooling (MAP)~\citep{zhai2022scaling,lee2019set}. The text-encoder uses a sentencepiece tokenizer~\citep{kudo2018sentencepiece} trained on the English-C4~\citep{raffel2020exploring} dataset, with a vocabulary size of $32,000$. We truncate all text captions to the first $64$ tokens. 
For most experiments, we use an rsqrt learning rate scheduler~\citep{zhai2022scaling}, with a peak learning-rate of $0.001$, and linear-warmup and linear-cooldown applied for $10\%$ of total steps. However, for some of our final method comparisons in~\cref{tab:full_eval}, we use a cosine learning rate scheduler~\citep{roth2024practitioner} with a linear-warmup applied for $10$\% of total steps and peak learning-rate of $0.001$.
By default, we use a filtering ratio of ${f}{=}{0.8}$ when using \ourmethod sampling, leading to a super-batch-size of ${B}{=}{163,840}$.
We additionally use an ACID sampling temperature of ${\tau}{=}{10}$ for all our experiments.
We sweep over ${\lambda}{=}{\{{0.5}{,}{1.0}{,}{2.0}\}}$ for finding the optimal loss-weight for the \textit{Softmax-Distillation} loss (\cref{eq:smax-distillation-loss}).
We use a weight decay of $0.0001$, gradient clipping to a maximum
norm of $1.0$, and the Adam optimizer with (${\beta_{1}}{=}{0.9}$, ${\beta_{2}}{=}{0.95}$). All our experiments are conducted with \texttt{big\_vision}~\citep{bigvision} using \texttt{jax}~\citep{jax2018github}.

\newpage
\section{About baselines and final ACED models}
\label{expansion-on-model-details}


In this section, we describe the exact architectural details of all the baselines and our ACED models in~\cref{tab:appendix-arch-details}.

\begin{table*}[h!]
    \centering
    \caption{\textbf{Architectural Details of baselines and \ourmethodwithkd-F* models.} For each of the baselines and our own \ourmethodwithkd models, we provide the exact image and text encoder architectures used, the image-resolution used for training, the patch-size for vision-transformer specific encoders, the text sequence-length, training dataset and total compute budget for training in terms of total samples seen.}
    \resizebox{0.99\linewidth}{!}{
        \begin{tabular}{l|C{1.2cm}|C{1.2cm}|C{4.0cm}|C{2.3cm}C{2.3cm}C{1.6cm}C{1.6cm}C{1.4cm}}
            \toprule[1.5pt]
            \multirow{1}{*}{\textbf{Method}}
            & {\centering\textbf{Samples Seen}}
            & {\centering\textbf{Infer. GFlops}}
            & {\centering\textbf{Pretraining Dataset}}
            & {\centering\textbf{Image Encoder}}
            & {\centering\textbf{Text Encoder}}
            & {\centering\textbf{Image\\Resolution}}
            & {\centering\textbf{Image\\Patch Size}}
            & {\centering\textbf{Text\\Seq. Len.}} \\

             \midrule[1.25pt]

             \textcolor{gray}{DatologyAI-cls-S/32} & \textcolor{gray}{2.0B} & \textcolor{gray}{2.83}  & \textcolor{gray}{Datology-Proprietary}
              & \textcolor{gray}{ViT-S/32} & \textcolor{gray}{BERT-small} & \textcolor{gray}{224} & \textcolor{gray}{32} & \textcolor{gray}{77} \\

             \textcolor{gray}{DatologyAI-ret-S/32} & \textcolor{gray}{2.0B} & \textcolor{gray}{2.83}  & \textcolor{gray}{Datology-Proprietary}
              & \textcolor{gray}{ViT-S/32} & \textcolor{gray}{BERT-small} & \textcolor{gray}{224} & \textcolor{gray}{32} & \textcolor{gray}{77} \\

             TinyCLIP-RN30M & 15.2B** & 6.93 & LAION-400M & RN-30M & Custom & 224 & (-) & 77  \\

             TinyCLIP-45M/32 & 15.8B** & 3.70 & LAION+YFCC-400M & ViT-65M/32 & Custom  & 224 & 32 & 77  \\

             TinyCLIP-63M/32 & 15.8B** & 5.65 & LAION+YFCC-400M & ViT-63M/32 & Custom  & 224 & 32 & 77  \\

             MobileCLIP-S0 & 13B* & 3.70 &
             DataCompDR-1B & MCi0 & MCt & 256 & (-) & 77  \\

             \rowcolor{DnCBG}{\textbf{{ACED}-F0}} & 13B & {3.30} & DataComp-1B & ViT-S/32 & BERT-small & 256
             & 32 & 64 \\  

             \midrule

             \textcolor{gray}{DatologyAI-cls-B/32} & \textcolor{gray}{5.1B} & \textcolor{gray}{7.39}  & \textcolor{gray}{Datology-Proprietary}
              & \textcolor{gray}{ViT-B/32} & \textcolor{gray}{BERT-base} & \textcolor{gray}{224} & \textcolor{gray}{32} & \textcolor{gray}{77} \\

             \textcolor{gray}{DatologyAI-ret-B/32} & \textcolor{gray}{5.1B} & \textcolor{gray}{7.39}  & \textcolor{gray}{Datology-Proprietary}
              & \textcolor{gray}{ViT-B/32} & \textcolor{gray}{BERT-base} & \textcolor{gray}{224} & \textcolor{gray}{32} & \textcolor{gray}{77} \\

             CLIP-KD-RN50 & 0.5B & 9.09 & CC-3M+CC-12M
              & RN-50 & BERT-base & 224 & (-) & 77  \\
              
             OpenAI-RN50 & 13B & 9.09 & 
             OpenAI-WIT & RN-50 & BERT-base & 224 & (-) & 77 \\

             OpenAI-CLIP-B/32 & 13B & 7.39 & OpenAI-WIT
              & ViT-B/32 & BERT-base & 224 & 32 & 77  \\

             LAION-CLIP-B/32 & 34B & 7.39 & LAION-2B
              & ViT-B/32 & BERT-base & 224 & 32 & 77 \\

             DataComp-CLIP-B/32 & 13B & 7.39 & DataComp-1B
              & ViT-B/32 & BERT-base & 224 & 32 & 77  \\

             MetaCLIP-CLIP-B/32 & 13B & 7.39 & MetaCLIP-2B
              & ViT-B/32 & BERT-base & 224 & 32 & 77  \\

             CLIP-CID-B/32 & 7.2B & 7.39 & LAION-225M
              & ViT-B/32 & BERT-base & 224 & 32 & 77 \\

            TinyCLIP-39M/16 & 20B** & 9.48  & YFCC-15M & ViT-39M/16 & Custom  & 224  & 16 & 77   \\

             MobileCLIP-S1 & 13B* & 7.64 &
             DataCompDR-1B & MCi1 & BERT-base  & 256  & (-) & 77  \\

             \rowcolor{DnCBG}{\textbf{ACED-F1}} & 13B & {7.14} & DataComp-1B & ViT-B/32 & BERT-small & 256 & 32 & 64 \\

             \midrule

             OpenAI-RN101 & 13B & 12.75 & 
             OpenAI-WIT & RN-101 & BERT-base & 224 & (-) & 77 \\

             MobileCLIP-S2 & 13B* & 10.81 & DataCompDR-1B & MCi2 & BERT-base  & 256  & (-) & 77  \\

          \rowcolor{DnCBG}{\textbf{ACED-F2}} & 13B & {10.29} & DataComp-1B & ViT-B/24 & BERT-small & 240 & 24 & 64 \\

            \bottomrule[1.5pt]
        \end{tabular}}
    \vspace{-5pt}
    \label{tab:appendix-arch-details}
    \vspace{-5pt}
\end{table*}


\newpage
\section{Comparison with other batch selection methods}
In this section, we compare our ACID method with other online batch selection methods in the literature as outlined in~\cref{sec:relwork}. For a fair comparison, we re-implement four batch-selection methods under our setting, namely, Bad-Students~\citep{evans2023bad}, Selective-Backprop~\citep{jiang2019accelerating}, RHO-Loss~\citep{mindermann2022prioritized} and JEST~\citep{evans2024data}. For this experiment, we pretrain SigLIP models on DataComp-1B~\citep{gadre2024datacomp} for 3B samples seen. For the reference models required by RHO-Loss, JEST and ACID, we use our pretrained WebLI-C++ reference.From~\cref{tab:otherbatchselbaselines}, we observe that our ACID method outperforms all the other batch-selection methods by large margins ($1.4\%$ better than JEST and $3.2\%$ better than RHO-loss).

\begin{table}[!h]
    \centering
    \begin{tabular}{l|cc|c}
            \toprule
            \textbf{Method} & \textbf{IN-val} & \textbf{COCO} & \textbf{27-Avg}\\
            \midrule    
            \textcolor{gray}{\textbf{IID (baseline)}} & \textcolor{gray}{63.6} & \textcolor{gray}{42.4} &\textcolor{gray}{60.1} \\
            \textcolor{gray}{\textbf{Softmax-KD}} & \textcolor{gray}{66.1} & \textcolor{gray}{47.3} & \textcolor{gray}{62.0} \\
            \midrule
            \textbf{{{{Bad-Students~\citep{evans2023bad}}}}} & 60.9 & 49.0 & 57.8   \\
            \textbf{{{{Sel-BP~\citep{jiang2019accelerating}}}}} & 63.5 & 42.7 & 60.2\\
            \textbf{{{{RHO-loss~\citep{mindermann2022prioritized}}}}} & 65.9 & 49.4 & 62.6  \\
            \textbf{{{{JEST~\citep{evans2024data}}}}} & 68.7 \underline{} & 53.4 & 64.4  \\
            \midrule
            \rowcolor{DnCBG}{\textbf{ACID}} & \textbf{71.0}  & \textbf{53.6} & \textbf{65.8} \\
            \bottomrule
    \end{tabular}
    \caption{\textbf{ACID outperforms all other online batch selection methods.}}
    \label{tab:otherbatchselbaselines}
\end{table}

\newpage
\section{Additional Experiments,  Ablations and Results}


In this section, we provide some additional ablations and more detailed results, augmenting those present in the main paper. We further also include additional baseline comparisons with proprietary models.

\subsection{ACIDistill vs. IIDistill scaling}

\label{sec-acidistill}
\begin{figure*}[h]
    \centering
    \includegraphics[scale=0.5]{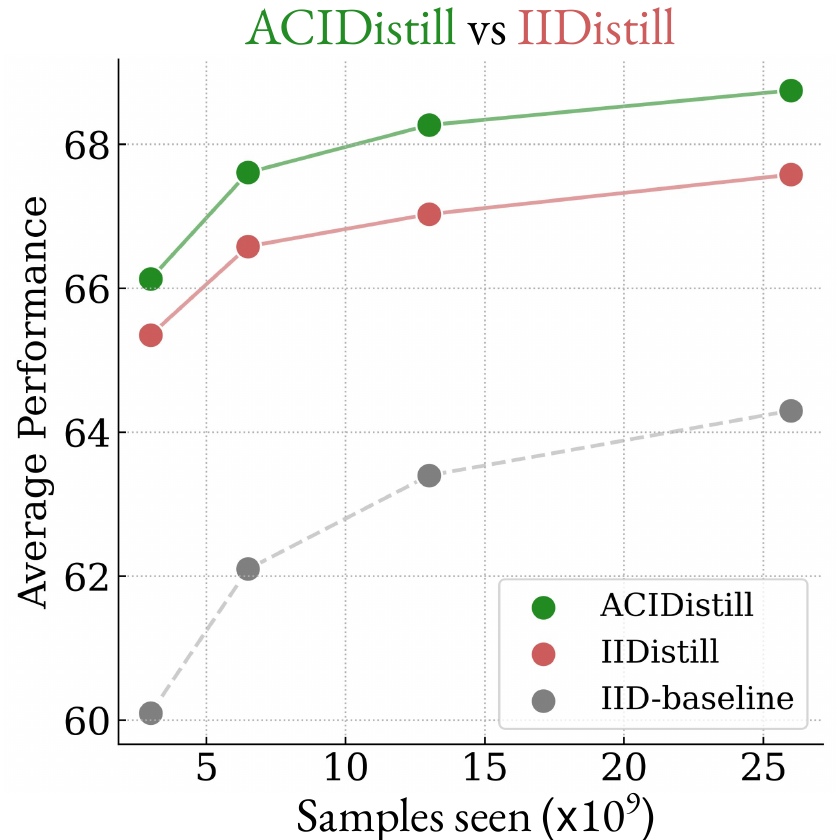}
    \caption{\textbf{How to combine ACID and KD in ACED?} The optimal scalable strategy for combining \ourmethod and \textit{Softmax-Distillation} is the \textit{ACIDistill} method---where we apply both the contrastive and distillation losses on the \ourmethod batch---this is both more performant and training-time efficient than the \textit{IIDistill} scheme.}
    \label{fig:appx_acid_kd_combine}
\end{figure*}
\subsection{Softmax vs Sigmoid Pretraining}
We have used SigLIP (sigmoid) pretraining for all our main results because of it's strong performance as a baseline. Here we show that the results are similar with CLIP (softmax) pretraining as well. Overall, the sigmoid variant is more scalable.

\begin{figure*}[h]
\centering
\begin{subfigure}[b]{0.4\textwidth}
    \centering
    \includegraphics[scale=0.4]{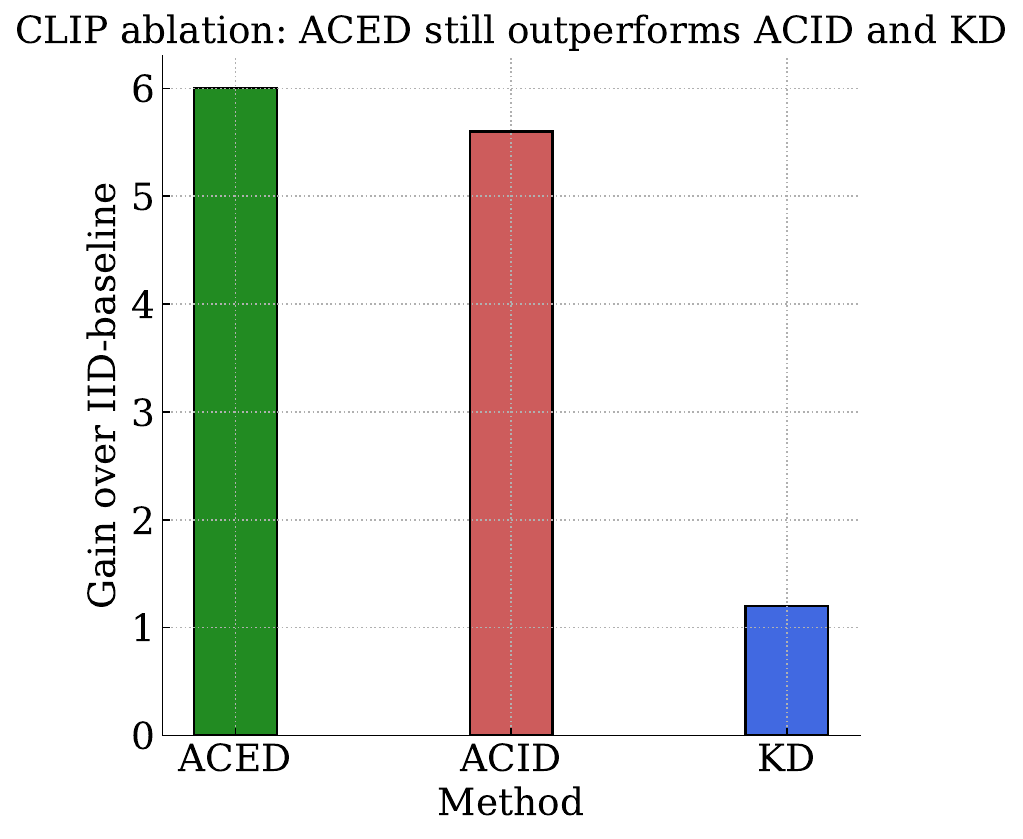}
    \label{fig:appx-clip-ablation}
\end{subfigure}\hspace{1cm}
\begin{subfigure}[b]{0.4\textwidth}
    \centering
    \includegraphics[scale=0.4]{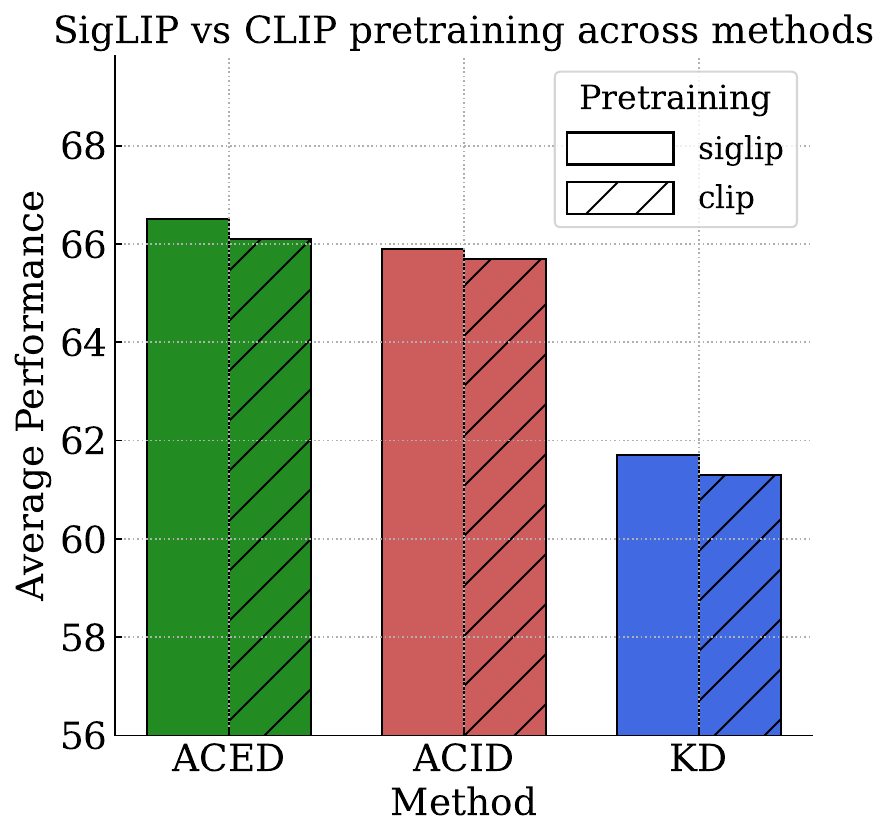}
    \label{fig:appx-clip-vs-siglip}
\end{subfigure}
\vspace{-5pt}
\caption{\textbf{CLIP vs SigLIP pretraining.} \textit{(left)} Our \textit{ACED} method when applied with CLIP pretraining instead of SigLIP, also further improves over both our \textit{ACID} and \textit{Softmax-KD} approaches. This showcases our methods' generality across pretraining objectives. \textit{(right)} We compare all our methods across SigLIP and CLIP pretraining, and we observe that SigLIP pretraining clearly outperforms the CLIP objective across all the methods, justifying our choice of using it for all our final results.}
\label{appendix:clip-vs-siglip-ablation-appendix}
\end{figure*}

\subsection{ACID vs KD as we scale compute}
\begin{figure*}[h]
    \centering
    \includegraphics[scale=0.5]{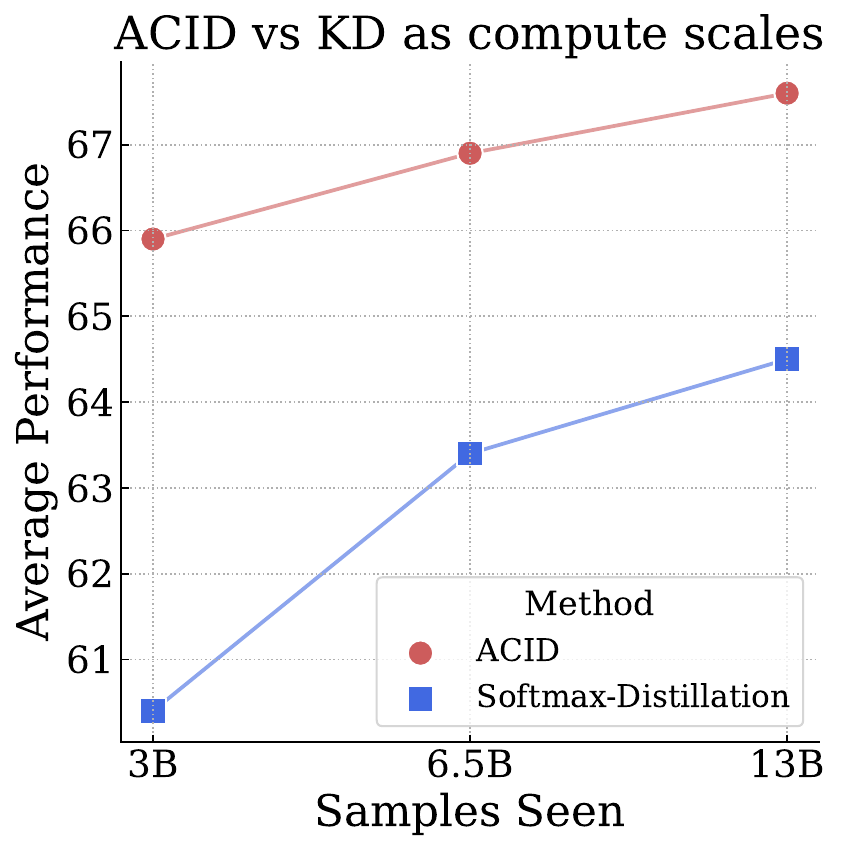}
    \caption{\textbf{ACID outperforms Softmax-Distillation across training compute budgets.}}
    \label{fig:appx-acid-kd-compute-budget}
\end{figure*}

In~\cref{sec:selectdist-outperform}, we demonstrated that our \textbf{\textit{ACID}} outperforms distillation methods across a variety of data-, student-size-, and method-configurations. However, all these results were at the 3B samples seen scale. Here, we compare \textit{ACID} and \textit{Softmax-Distillation} as we increase the training compute budget to $6.5$B and $13$B samples seen scale.~\cref{fig:appx-acid-kd-compute-budget} depicts that as we scale up the compute budget, \textit{ACID} still strongly outperforms \textit{Softmax-Distillation}, further signifying the scalability of our method.

\subsection{Full Detailed Results across all 27 \textit{StableEval} Evaluations}

\begin{table*}[h]
    \footnotesize
    \centering
    \setlength\tabcolsep{2pt}
    \caption{\textbf{Full Detailed Per-Dataset Results of ACED models on the 27 \textit{StableEval} Evaluations.}}
    \resizebox{0.99\linewidth}{!}{
    \begin{tabular}{c|ccccccccccccccccccccccccccc|c}
    \toprule
         & \rotatebox{90}{\textbf{FGVC-Aircrafts}} & \rotatebox{90}{\textbf{Oxford-Flowers-102}} & \rotatebox{90}{\textbf{Oxford-IIIT-Pets}} & \rotatebox{90}{\textbf{Stanford Cars}} & \rotatebox{90}{\textbf{Food-101}} & \rotatebox{90}{\textbf{Caltech-101}} & \rotatebox{90}{\textbf{CIFAR-10}} & \rotatebox{90}{\textbf{CIFAR-100}} & \rotatebox{90}{\textbf{Pascal VOC 2007}} & \rotatebox{90}{\textbf{STL-10}} & \rotatebox{90}{\textbf{SUN-397}} & \rotatebox{90}{\textbf{GeoDE}} & \rotatebox{90}{\textbf{RESISC45}} & \rotatebox{90}{\textbf{FMoW}} & \rotatebox{90}{\textbf{ImageNet-V2}} & \rotatebox{90}{\textbf{ImageNet-A}} & \rotatebox{90}{\textbf{ImageNet-R}} & \rotatebox{90}{\textbf{ObjectNet}} & \rotatebox{90}{\textbf{ImageNet-Val}} & \rotatebox{90}{\textbf{ImageNet-Sketch}} & \rotatebox{90}{\textbf{DTD}} & \rotatebox{90}{\textbf{DollarStreet}} & \rotatebox{90}{\textbf{Country211}} & \rotatebox{90}{\textbf{Flickr30k I2T}} & \rotatebox{90}{\textbf{Flickr30k T2I}} & \rotatebox{90}{\textbf{COCO I2T}} & \rotatebox{90}{\textbf{COCO T2I}} & \rotatebox{90}{\textbf{Average (27)}}\\
    \midrule
    \rotatebox{0}{\ourmethodwithkd-F0} & 18.75 & 73.41 & 89.13 & 79.23 & 85.41 & 84.40 & 93.88 & 74.38 & 83.60 & 97.28 & 69.12 & 86.94 & 64.94 & 16.46 & 61.21 & 33.05 & 79.15 & 51.05 & 68.45 & 53.37 & 45.69 & 43.73 & 15.09 & 87.60 & 71.40 & 60.80 & 41.23 &  64.0 \\
    \rotatebox{0}{{\ourmethodwithkd-F1}} & 26.94 & 79.59 & 91.31 & 83.32 & 89.96 & 85.24 & 96.69 & 81.68 & 84.39 & 98.78 & 73.13 & 90.49 & 69.49 & 23.09 & 67.80 & 53.35 & 87.93 & 60.24 & 74.92 & 61.55 & 51.54 & 48.49 & 20.28 & 90.30 & 77.92 & 64.96 & 47.27 & 69.7 \\
    \rotatebox{0}{\ourmethodwithkd-F2} & 27.00 & 79.41 & 92.29 & 86.48 & 91.12 & 83.99 & 96.03 & 82.86 & 85.37 & 98.85 & 74.06 & 91.19 & 68.84 & 24.21 & 70.03 & 58.64 & 90.14 & 63.87 & 76.90 & 63.67 & 50.21 & 48.42 & 22.10 & 91.10 & 79.46 & 66.92 & 49.69 & 70.9 \\
    \bottomrule
    \end{tabular}}
    \label{tab:appendix-full-27-evals}
\end{table*}

\newpage
\subsection{Hyperparameter Sensitivity in ACID}

\begin{figure*}[h]
\centering
\begin{subfigure}[b]{0.4\textwidth}
    \centering
    \includegraphics[scale=0.4]{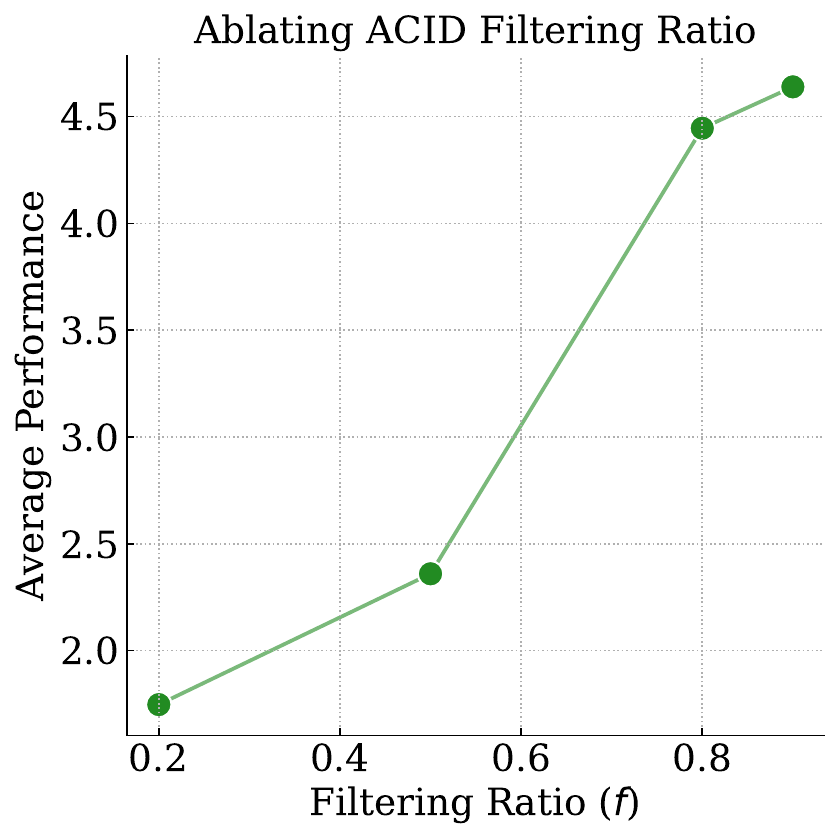}
    \label{fig:appx-f-ablation}
\end{subfigure}\hspace{1cm}
\begin{subfigure}[b]{0.4\textwidth}
    \centering
    \includegraphics[scale=0.4]{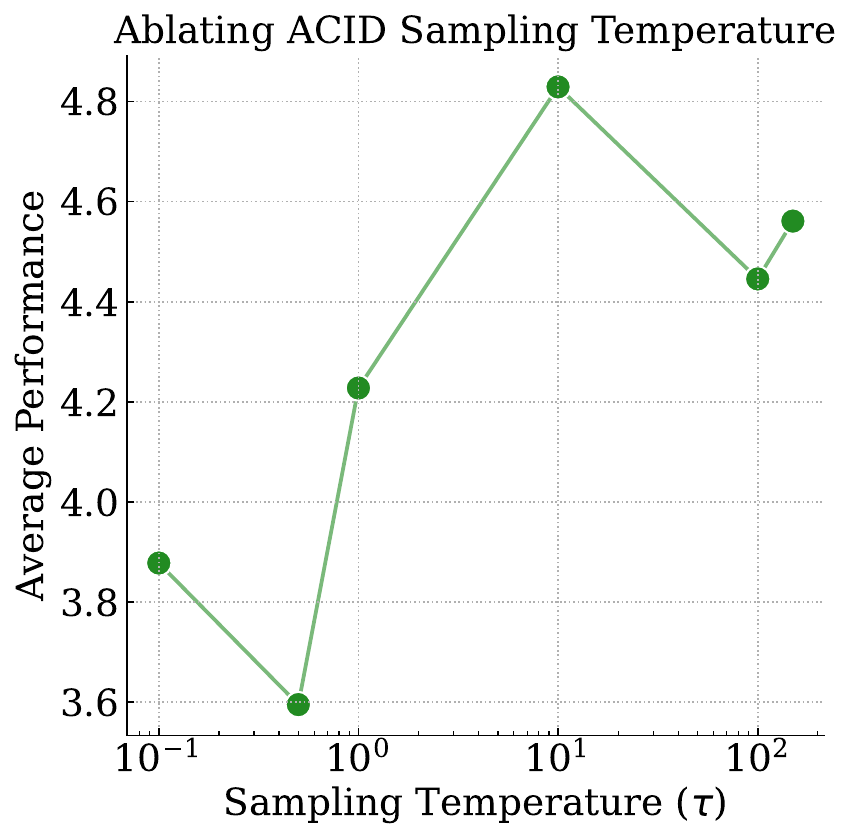}
    \label{fig:appx-tau-ablation}
\end{subfigure}
\vspace{-5pt}
\caption{\textbf{ACID hyperparameters.} \textit{(left)} We observe that as we keep increasing the filtering ratio, we continue to see improved performance from ${f}{=}{0.2}$ to ${f}{=}{0.8}$. However, note that these improvements saturate at very high filtering ratios (${f}{=}{0.9}$) due to very aggressive filtering which might lead to insufficient coverage of the entire data distribution. \textit{(right)} We find a sampling temperature ${\tau}{=}{10}$ to be optimal across the range of sampling temperatures we tested, trading-off between deterministic top-k sampling (at very high temperatures) vs random sampling (at very low temperatures).
 }
\label{appendix:sampling-temp-and-filtering-ratio-ablations}
\end{figure*}

\newpage
\section{Extended Related Works}
\noindent\textbf{Multimodal Data Curation.} Recent works have emphasised the importance of data quality for pretraining multimodal models~\citep{nguyen2022quality,li2024datacomp,gadre2024datacomp,udandarao2024no,fang2022data,mayilvahanan2023does}. Canonical methods for curating high-quality training data generally involve static offline curation include removing noisy samples~\citep{jia2021scaling,changpinyo2021conceptual,wang2023too,sorscher2022beyond,abbas2023semdedup,cao2023less,maini2023t, xu2023demystifying,abbas2024effective, gadre2024datacomp}, rebalancing concept distributions~\citep{oquab2023dinov2,xu2023demystifying,abbas2024effective}, improving quality of text captions~\citep{li2024if,yu2024capsfusion,lai2023scarcity,fan2024improving,nguyen2024improving,li2022blip,li2023blip,yang2023alip,zhang2023compress,nguyen2024multilingual}, and using pretrained data-selector models for filtering samples with low image-text alignment~\citep{mahmoud2024sieve,wang2024variance,fang2023data,wang2024cliploss,schuhmann2021laion,schuhmann2022laion,kim2024hype,yu2023devil,wang2024finetuned}. Specifically, it has been shown that offline curation of noisy web-scale data can result in large pretraining efficiency gains \citep{sorscher2022beyond,jia2021scaling,changpinyo2021conceptual,wang2023too,abbas2023semdedup,cao2023less,maini2023t, xu2023demystifying,abbas2024effective, mahmoud2024sieve,wang2024variance,fang2023data,wang2024cliploss,schuhmann2021laion,kim2024hype,yu2023devil,wang2024finetuned,vo2024automatic}.

However, such static offline curation methods that pre-filter data do not take into account the training dynamics of the current learner model, and hence can suffer at larger scales~\citep{goyal2024scaling}. Some prior works tackle this by introducing data selection criteria that account for the current state of the learner---~\citet{loshchilov2015online} proposed \textit{online batch selection}, that at each step selects training samples that have the largest learner loss. Further works extended upon this idea by exploring different sample selection criteria, all based on the current learner state~\citep{ioannou2023online,kumar2010self,fan2016neural,katharopoulos2018not,joseph2019submodular,song2020carpe,jiang2019accelerating,wang2024efficienttrain,zhou2024multi,schaul2015prioritized,xu2023cit,toneva2018empirical,feldman2020does,maini2022characterizing,sachidananda2023global}.
Further,~\citet{mindermann2022prioritized} introduced the RHO-Loss that considers both current learner state and a pretrained data-selector (reference) model. Further works extended this criterion (termed \textit{learnability scoring}) and scaled it to foundation model training~\citep{evans2023bad,evans2024data,brandfonbrener2024color,hong2024diversified,fan2023irreducible,deng2023towards}.
A key underlying goal of almost all of these prior data curation methods is to improve training efficiency by reducing the number of samples required for pretraining. Owing to this push for training efficiency, most pretrained reference models that are used as \textit{data selectors are typically smaller than the learner models they are used to train}~\citep{evans2023bad,evans2024data,fang2023data}. In fact,~\citet{fang2023data,yu2023devil,gadre2024datacomp} all showed that increasing the reference model size might even be detrimental for training a good learner model.


In this work, we show for the first time that \textit{larger reference models can indeed be used as strong data selectors}, and showcase the conditions under which this simple active data-curation method can be used as an effective distillation strategy for training smaller learner models. Our experiments demonstrate that this can in-fact even outperform standard knowledge distillation strategies that are the most popular methods for compressing big models into smaller, more efficient ones.

\noindent\textbf{Knowledge Distillation.} First introduced by~\citet{buciluǎ2006model} and further popularized by~\citet{hinton2015distilling,ba2014deep}, knowledge distillation (KD) is a classic technique for transferring knowledge from a larger model (\textit{teacher}) to another smaller one (\textit{student}), by optimizing the student to match certain outputs (logits, features, intermediate activations etc.) of the teacher model. It has been extensively used for compressing large models into smaller, deployable ones in unimodal tasks like image-classification~\citep{beyer2022knowledge,cho2019efficacy,vemulapalliknowledge,wang2020knowledge,nix2023hard,tian2019contrastive,heo2019knowledge,romero2014fitnets,fang2021mosaicking,chen2019data,whatmakesgooddistillation} and language representation learning~\citep{xu2024survey,agarwal2024policy,sanh2019distilbert,kim2016sequence,lin2020autoregressive,hahn2019self,tan2023gkd}. Further works have extended KD to use multiple teacher-ensembles~\citep{shen2020meal,chebotar2016distilling,you2017learning,stanton2021does,malinin2019ensemble,faghri2023reinforce,zuchniak2023multi,sariyildiz2024unic}, different distillation training objectives~\citep{ji2021show,zhao2024no,xie2020self,tarvainen2017mean,touvron2021training,ranzinger2024radio,li2024promptkd}, and progressive multi-stage training schemes~\citep{andonian2022robust,han2024amd,zhao2024multistage,zhang2024progressive,liang2024module}. 
See~\citet{gou2021knowledge} for a comprehensive survey of KD methods across a range of practical unimodal settings.

However, KD methods in the multimodal foundation model regime are
underexplored. Some initial works~\citep{wang2022efficientvlm,liu2021kd,fang2021compressing,wang2021distilled,croitoru2021teachtext} proposed strategies for efficiently compressing a multimodal teacher for captioning, visual question-answering and video retrieval tasks.~\citet{sameni2024building} introduced SF-CLIP, a method for improving CLIP pretraining via masked distillation, while~\citet{vasu2024mobileclip} proposed MobileCLIP, exploring downscaling CLIP models for mobile-deployment by using a combination of multi-teacher contrastive-KD, synthetic captions, and data-augmentations.~\citet{wu2023tinyclip} further proposed TinyCLIP---a weight inheritance method combined with an affinity-mimicking strategy for multimodal KD to yield tiny CLIP models.~\citet{yang2023clip} conducted an extensive empirical study (CLIP-KD) into the different objective functions for effectively performing distillation of CLIP models, across different scales.
Finally, CLIP-CID~\citep{yang2024clip} uses an image semantic balancing strategy coupled with cluster-instance discrimination for better teacher-to-student knowledge transfer during the KD process.
We compare against these methods as baselines for our experimental results in~\cref{sec:experiments}.

\noindent\textbf{Accelerating Knowledge Distillation with Data Selection.} There have been prior works attempting to make KD-based pretraining more efficient~\citep{shen2022fast,shen2023ferkd,yun2021re}.
Some works~\citep{wang2020neural,lan2024improve,xu2023computation,baykal2022robust} have investigated accelerating vanilla KD using active learning in small-scale classification tasks. However, such approaches require a costly iterative process, involving synthetic generation, followed by active sample selection to produce pseudo-labels from a teacher model, thereby limiting their scalability. 
Another line of work studies data-selection methods for improving KD, typically using uncertainty-based data, logit and feature selection~\citep{rao2023dynamic,roth2023fantastic,wang2023improved,wang2022multimodal,lin2022efficient,zhou2023adads,li2021dynamic,he2022knowledge,wang2024cascade}, contextual retrieval and sample augmentation from a large data pool~\citep{liu2022rethinking,jiao2019tinybert,liang2020mixkd,ge2024training,zhang2023reaugkd,radenovic2023filtering,rawat2024little}, or influence-function based sample selection~\citep{lan2024improve,ye2022progen}. 
Contrary to these works,~\citet{beyer2022knowledge} and~\citet{hao2024revisit} suggest that vanilla knowledge distillation provides optimal gains in the ``infinite-data regimes''.
All these prior works however operate primarily in the unimodal image or text classification regime, and none has been scaled up to multimodal foundation model training.
We showcase, for the first time, that simple data selection using online batch selection outperforms standard KD for pretraining multimodal models. We further study the optimal strategies for combining vanilla KD and active data curation 
to best leverage their complementary strengths.


\newpage
\section{Discussion on training cost vs baselines}

In this section, we describe in detail the training costs required by ACID compared to other methods. We first define $F_{I}$ as the FLOPs-per-iteration of a forward pass of the image encoder of the student model. Similarly, we define $F_{T}$ as the FLOPs-per-iteration of a forward pass through the student text encoder. We do not consider the cost of the forward passes of teacher/reference models because we can cache their embeddings, as proposed in prior work~\citep{vasu2024mobileclip,vasu2024mobileclip}. 

\noindent Given this, we compute the total FLOPs per normal IID iteration is 3($F_I$ + $F_T$). After caching reference embeddings, scoring the super-batch with the student model adds 4($F_I$ + $F_T$) for a filtering ratio of $0.8$, which gives a total FLOPs / iteration of 7($F_I$ + $F_T$) (7/3x overhead compared to IID training). 

\noindent In~\cref{fig:appx-acid-kd-compute-budget}, we show that the ACID method trained for 3B examples outperforms Softmax-KD training at 13B examples. Even with the 7/3x overhead, the absolute gains of using ACID are significant compared with additional IID and Softmax-KD training. 
Further, the main SoTA competition, MobileCLIP~\citep{vasu2024mobileclip} has additional forward and backward passes due to an additional synthetic caption batch. This is an overhead of 3($F_I$ + $F_T$) - $F_I$ because the initial image forward-pass can be cached for the second batch. This gives a total FLOPs per iteration of 6($F_I$ + $F_T$) - $F_I$. If we compare for example, MobileCLIP-S0 (3.70 inference FLOPs) to ACED-F0 (3.30 inference FLOPs), the training per iteration of MobileCLIP-S0 = 6(2.39 + 1.32) - 2.39 = 19.81 FLOPs and ACED-F0 =  7(3.30) = 23.1 FLOPs. Thus ACED incurs an approx. 15\% training overhead compared with MobileCLIP. However, it is worth noting that the methods proposed in~\citet{evans2023bad} for flexible resolution scoring can be used to bring this training budget of ACED down drastically to well below that of MobileCLIP, with little loss in performance. We did not implement this as it has been shown in that prior work. Additionally, although MobileCLIP may have a slight training efficiency, their requirement for generating synthetic captions on new data is far more compute intensive than generating embeddings via our reference-model. Finally, we highlight that in general the main goal of our work (and others) \textit{is to maximize performance at given inference budgets} as it is generally assumed that the training cost of efficient models will be amortized over model lifetime in use.

\newpage
\section{Discussion}
\label{app:disc}

Model-based active learning and knowledge-distillation are separate techniques that have traditionally targeted two very different problems. While active learning via online batch selection has focused on improving performance and efficiency of large-scale foundation model pretraining, knowledge-distillation methods seek to achieve highly inference-efficient models by transfer of knowledge from these larger foundation models. In this work, we show theoretically that in fact, active data selection can be cast as a form of implicit knowledge-distillation where the target distribution is now a product of reference (teacher) model probabilities and real labels. With this insight, we develop \ourmethod, a powerful method for distilling efficient contrastive multi-modal encoders from larger reference models via online joint-example selection \citep{evans2024data}. Notably, this method is a significant and initially counterintuitive departure from traditional active curation paradigms~\cite{mindermann2022prioritized,evans2023bad} which typically seek reference models that are significantly cheaper in compute compared to the student.

We empirically validate that indeed \ourmethod is a strong form of distillation that strictly outperforms traditional forms of knowledge-distillation in training contrastive VLMs. Given the different form of implicit distillation objective in \ourmethod, we further demonstrate that this is complementary with traditional softmax-based KD, arriving at a final method, \ourmethodwithkd, which combines the benefits of each. Using \ourmethod we effectively distill models that achieve stronger zero-shot classification and image-text retrieval with cheaper inference FLOPs than prior SoTA methods.

\subsection{Limitations}

While we see our work as a novel, simple, and scalable paradigm for effective distillation of efficient models, our results are limited in scope to contrastive training of VLMs. Knowledge-distillation can in theory be applied to many problems such as supervised image classification~\citep{kolesnikov2020big}, self-supervised learning~\citep{chen2020simple,grill2020bootstrap}, etc. and it remains to be seen whether our results can be transferred to these domains. Furthermore, while we have shown that we can distill SoTA models that are efficient on a theoretical FLOPs basis, it remains to be seen whether our method can achieve SoTA results when constrained by device latency as is necessary for many edge deployments. We leave it to future work to benchmark our method with SoTA low-latency architectures like FastVIT~\citep{vasu2023fastvit} or MobileNet-V4~\citep{qin2025mobilenetv4}.

\end{document}